\documentclass{article}

\usepackage{microtype}
\usepackage{graphicx}
\usepackage{subcaption}
\usepackage{booktabs,multirow,multicol} 
\usepackage{amsthm,amssymb,thmtools} 
\usepackage{amsmath}
\usepackage{mathtools}
\usepackage{relsize} 
\usepackage{xcolor}
\usepackage{paralist}
\usepackage{comment}
\usepackage{placeins}

\usepackage[accepted]{icml2021}
\usepackage{hyperref} 
\hypersetup{hidelinks}

\usepackage{cleveref}

\newcommand{\indep}{\perp \!\!\! \perp}

\usepackage{layouts}
\usepackage{wrapfig}
\usepackage{enumitem}

\usepackage{amsthm}
\usepackage{thmtools}
\usepackage{thm-restate}

\declaretheorem{definition}

\icmltitlerunning{Equivariant Learning of Stochastic Fields:  Gaussian Processes and Steerable Conditional Neural Processes}
\newcommand{\Reals}{\mathbb{R}} 
\newcommand{\Ind}{\text{Ind}} 
\newcommand{\eukl}[2]{\langle #1, #2 \rangle}

\newcommand{\GL}{\text{GL}}

\newcommand{\reg}{\text{reg}}
\newcommand{\tin}{\text{in}}
\newcommand{\tout}{\text{out}}

\newcommand{\Id}{\text{Id}}

\newcommand{\tforall}{\quad\text{for all }}

\usepackage{amsmath} 
\usepackage{amsfonts}
\usepackage{amssymb}
\usepackage{mathtools}

\def\grp{{G}}
\def\hrp{{H}}

\def\op{\cdot}

\def\inv{^{-1}}

\def\x{\mathbf{x}}
\def\y{\mathbf{y}}

\def\RR{\mathbb{R}}

\def\ZZ{\mathcal{Z}}

\DeclareMathOperator*{\vect}{vec}

\definecolor{mydarkblue}{rgb}{0,0.08,0.45}
\hypersetup{ 
    colorlinks=true, 
    linkcolor={blue!50!black},
    citecolor={blue!50!black}, 
    urlcolor={blue!80!black} 
}

\begin{document}

\twocolumn[
\icmltitle{Equivariant Learning of Stochastic Fields: \\ Gaussian Processes and  Steerable Conditional Neural Processes}



\icmlsetsymbol{equal}{*}

\begin{icmlauthorlist}
\icmlauthor{Peter Holderrieth}{equal,A}
\icmlauthor{Michael Hutchinson}{equal,A}
\icmlauthor{Yee Whye Teh}{A,B}
\end{icmlauthorlist}

\icmlaffiliation{A}{University of Oxford, United Kingdom}
\icmlaffiliation{B}{DeepMind, United Kingdom}

\icmlcorrespondingauthor{Peter Holderrieth}{peter.holderrieth@new.ox.ac.uk}


\icmlkeywords{Machine Learning, ICML}

\vskip 0.3in
]



\printAffiliationsAndNotice{\icmlEqualContribution} 

\begin{abstract}
Motivated by objects such as electric fields or fluid streams, we study the problem of learning \emph{stochastic fields}, i.e.
stochastic processes whose samples are \emph{fields} like those occurring in physics and engineering. Considering general
transformations such as rotations and reflections, we show that spatial invariance of stochastic fields requires an
inference model to be equivariant. Leveraging recent advances from the equivariance literature, we study equivariance
in two classes of models. Firstly, we fully characterise equivariant Gaussian processes. Secondly, we introduce
Steerable Conditional Neural Processes (SteerCNPs), a new, fully equivariant member of the Neural Process family. In
experiments with Gaussian process vector fields, images, and real-world weather data, we observe that SteerCNPs significantly
improve the performance of previous models and equivariance leads to improvements in transfer learning tasks.
\end{abstract}

\section{Introduction}
\label{sec:introduction}
\begin{figure}
    \centering
    \includegraphics{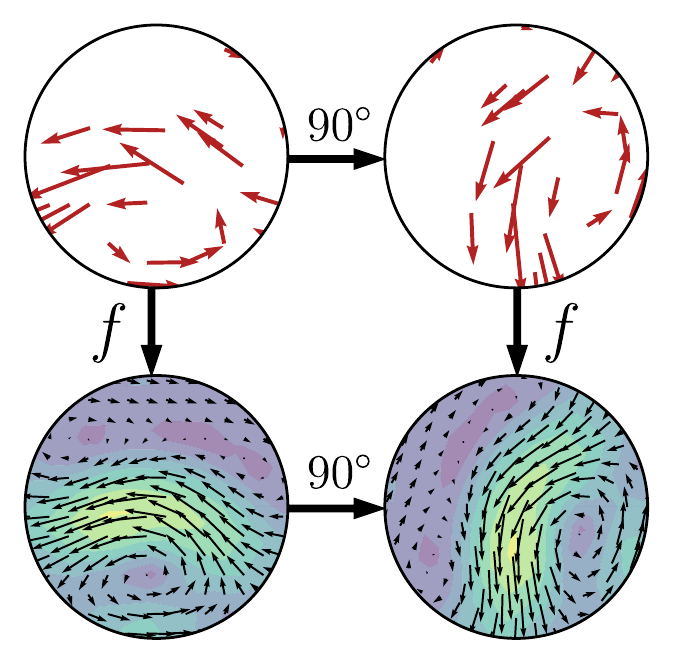}
    \caption{A schematic of equivariant learning of vector fields. The model $f$ predicts a vector field from a set of vectors (red arrows). It is equivariant to the action of rotations, here a 90$^\circ$ rotation.}
    \label{fig:eqv_demo}
\end{figure}
In physics and engineering, fields are objects which assign a physical quantity to every point in space. They serve as a unifying concept for objects such as electro-magnetic fields, gravity fields, electric potentials or fluid dynamics and are therefore omnipresent in the natural sciences and their applications \citep{landau2013classical}. Our goal in this work is to be able to predict the value of a field everywhere given some finite set of observations (see $f$ in \cref{fig:eqv_demo}). We will be interested in cases where these fields are not fixed but are drawn from a random distribution, which we term \emph{stochastic fields}.

Viewing fields as usual mathematical functions, we can consider stochastic fields as stochastic processes. A well-known example of these are Gaussian processes (GPs) \citep{GP_book} which have been widely used in machine learning. More recently, Neural Processes (NPs) and their related models were introduced as an alternative to GPs which enable to learn a stochastic process from data leveraging the flexibility of neural networks \citep{CNPs,NPs,attNP}. 





When applying models to data from the natural sciences, it
is a logical step to integrate scientific knowledge about the
problem into these models. The physical principle of homogeneity of space states that all positions and orientations in space are equivalent, i.e. there is no canonical orientation or absolute position. A natural modelling assumption stemming from this is that the prior we place over a random field should be invariant, i.e. it should look the same from all positions and orientations. We will show that this implies that the posterior as a function of the observed points is \textit{equivariant}. 

\Cref{fig:eqv_demo} illustrates this equivariance. The input is a discrete set of vectors at certain points in space, the red arrows, and the predictions a continuous vector field. If the space is homogenous, we would expect the predictions of a model to rotate in the same way as we rotate the data, i.e. we expect the model to be equivariant.

Translation equivariance in Gaussian processes has long been studied via stationary kernels. We extend this notion to derive Gaussian processes that are equivariant to more general transformations such as rotations and reflections. Recent work has also shown how to build translation equivariance into NP models \citep{ConvCNP,GaussNPs}. We will build on this work to introduce a new member of the Neural Process family that has more general equivariance properties. We do this utilising recent developments in equivariant deep learning \citep{E2Steerable}. Imposing equivariance on deep learning models reduces the number of model parameters and has been shown to allow models to learn from data more efficiently \citep{G_CNNs,CyclicSymmetry,KondTrivGen}. We will show that the same rational applies to NP models.

More specifically, our main contributions are as follows:
\begin{enumerate}
\item We show that stochastic process models are equivariant if and only if the underlying prior is spatially invariant - giving a natural criteria when such models are useful.
\item We find sufficient and necessary constraints for a vector-valued Gaussian process over $\RR^n$ to be equivariant allowing the simple construction of equivariant GPs.
\item As a new, equivariant member of the Neural Process family, we present Steerable Conditional Neural Processes (SteerCNPs) and show that they outperform previous models on synthetic and real data.
\end{enumerate}

\section{Transforming Fields}
\label{sec:feature_maps}
We aim to build to a model which learns functions $F$ of the form $F:\Reals^n\to\Reals^d$. We call $F$ a \emph{(steerable) feature map} since we interpret $F$ geometrically as mapping $n$-dimensional coordinates $\mathbf{x}\in\Reals^n$ to some $d$-dimensional feature $F(\mathbf{x})$. As intuitively clear from \cref{fig:eqv_demo}, we should be able to rotate such a feature map as we do with an ordinary geographical map or an image. In this section, we make this rigorous using group theory (see \cref{sec:intro_to_groups} for a brief introduction).

In the following, let $E(n)$ be the group of isometries on $\Reals^n$. Let $T(n)$ be the group of translations of $\RR^n$ which can be identified with $\RR^n$. $T(n)$ acts from the left on $\RR^n$ via $t_{\mathbf{x}}(\mathbf{x}')=\mathbf{x}'+\mathbf{x}$ for all $\mathbf{x},\mathbf{x}'\in\Reals^n$. Let $O(n)$ be the group of $n \times n$ orthogonal matrices, acting from the left on $\RR^n$ by matrix multiplication. We write $SO(n)$ for the subgroup of rotations, i.e. elements $A\in O(n)$ with $\det A=1$. 

We describe all possible transformations of a feature map $F$ by a subgroup $G \subset E(n)$. We assume that $G$ is the semidirect product of $T(n)$ and a subgroup $H$ of $O(n)$, i.e. every $g\in G$ is a unique composition of a translation $t_{\mathbf{x}}$ and an orthogonal map $h\in H$: 
\begin{align}
	g=t_{\mathbf{x}}h
\end{align}
As common, we call $H$ the \emph{fiber group}. Depending on the inference problem, one would often pick $H=SO(n)$ or $H=O(n)$ (equivalently $G=SE(n),E(n)$). However, using finite subgroups $H$ can be more computationally efficient, and give better empirical results \citep{E2Steerable}. In particular, in dimension $n=2$ we use the Cyclic groups $C_m$, comprised of the rotations by $\frac{k2 \pi}{m}$ ($k=0,1,\dots,m-1$) and the dihedral group $D_m$ containing $C_m$ combined with reflections.

\begin{figure}[h]
    \centering
    \includegraphics[width=\columnwidth]{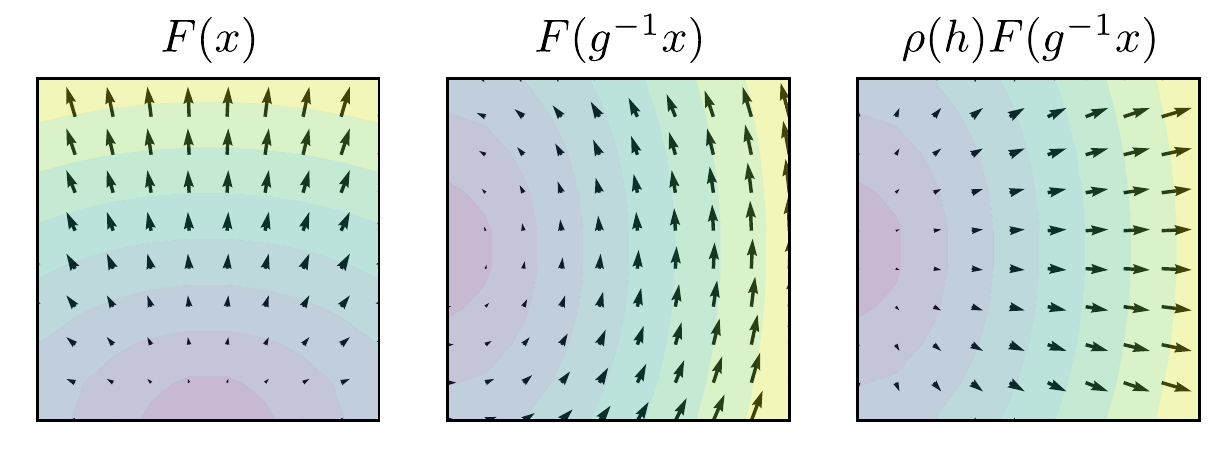}
    \caption{Demonstration of the transformation of vector fields on $\Reals^2$ under the action of $g = t_{\x'}h \in G=SE(2)=\Reals^2\rtimes SO(2)$. Color represents the norm of a vector at each point.}
    \label{fig:feature_field_transformation}
\end{figure}

To describe transformations of a feature map $F$ via a group $G$, we need a linear representation 
$\rho:H\to\GL(\Reals^d)$ of $H$ which we call \emph{fiber representation}. The action of $G$ on a steerable feature map $F$ is then defined as
\begin{equation}
    \label{eq:transforming_feature_map}
    g.F(\mathbf{x}) = \rho(h) F(g^{-1}\mathbf{x})
\end{equation}
where $g = t_{\mathbf{x}'} h\in G$. Figure \ref{fig:feature_field_transformation} demonstrates for vector fields why the transformation defined here is a sensible notion to consider. In group theory, this is called the induced representation of $H$ on $G$ denoted by $\Ind_{H}^{G}\rho$. 

In allusion to physics, we use the term \emph{(steerable) feature field} referring to the feature map $F:\Reals^n\to\Reals^d$ together with its corresponding law of transformation given by $\rho$ \citep{E2Steerable}. We write $\mathcal{F}_{\rho}$ for the space of these fields. Typical examples are:
\begin{enumerate}
	\item \textbf{Scalar fields} $F:\Reals^n\to\Reals$ have trivial fiber representation $\rho=\rho_{triv}$, i.e. $\rho(h) = \mathbf{1}$ for $h\in H$, such that
	\begin{align}
		g.F(\mathbf{x})=F(g^{-1}\mathbf{x})
	\end{align}
	Examples are greyscale images or temperature maps.
	\item \textbf{Vector fields} $F:\Reals^n\to\Reals^n$  have $\rho=\rho_{\text{Id}}$, i.e. $\rho(h) = h$  for $h\in H$, such that
	\begin{align}
	    \label{eq:vf_transformation}
		g.F(\mathbf{x})=hF(g^{-1}\mathbf{x}) 
	\end{align}
	Examples include electric fields or wind maps.
	\item \textbf{Stacked fields}: given fields $F_1,\dots,F_n$ with fiber representations $\rho_1,\dots,\rho_n$ we can stack them to $F=(F_1,\dots, F_n)$ with fiber representation as the direct sum $\rho=\rho_1\oplus\cdots\oplus\rho_n$. Examples include a combined wind and temperature map or RGB-images.
\end{enumerate}
Finally, for simplicity we assume that $\rho$ is an 
orthogonal representation, i.e. $\rho(h)\in O(d)$ for all $h\in H$. Since all fiber groups $H$ of interest are compact, this is not a restriction \citep{serreFiniteGroups}.

\begin{figure*}[h]
    \centering
     \begin{subfigure}[b]{0.58\textwidth}
         \centering
         \includegraphics[width=\textwidth]{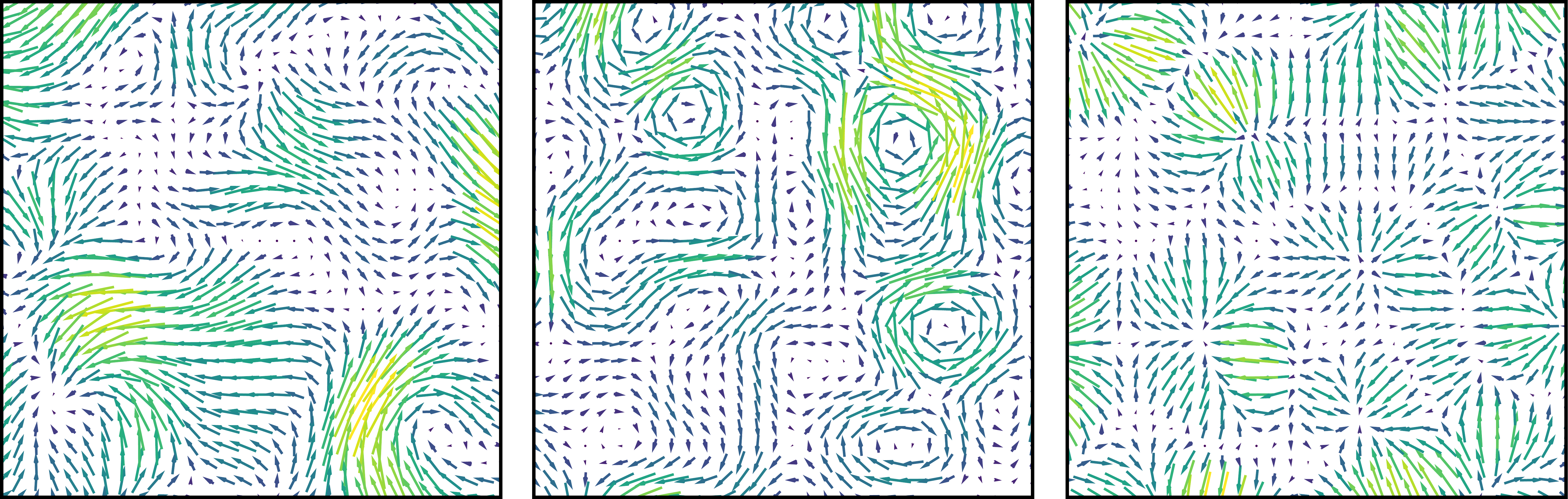}
         \caption{}
         \label{fig:gp_datasets}
     \end{subfigure}
     \hfill
     \begin{subfigure}[b]{0.38\textwidth}
         \centering
        \includegraphics[width=\textwidth]{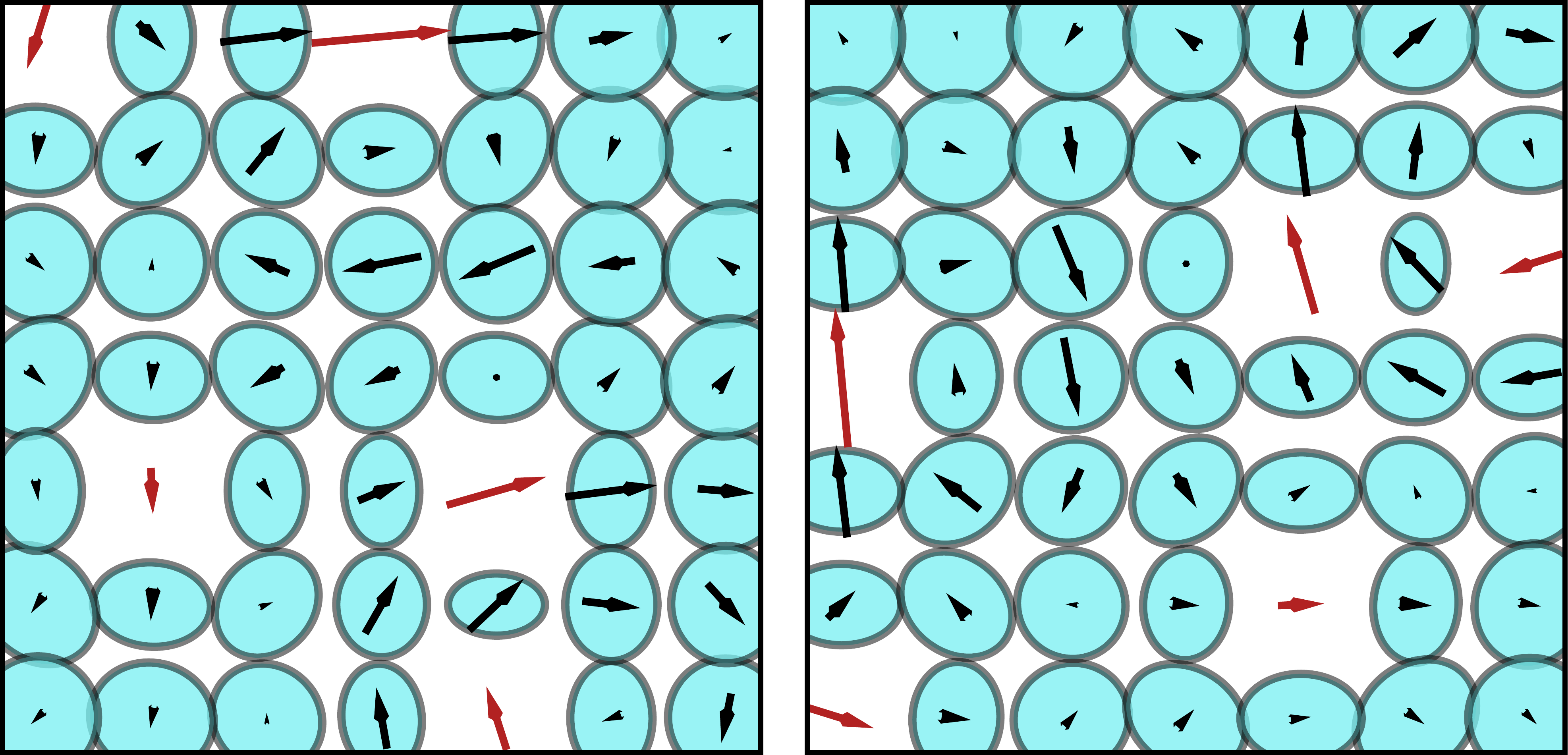}
         \caption{}
          \label{fig:gp_inference}
     \end{subfigure}
     \caption{(a) GP-samples of $2d$-vector fields from different $E(2)$-equivariant kernels. Left: RBF-kernel. Middle: divergence-free kernel. Right: curl-free kernel. (b) Illustration of \cref{thm:equivgp}. Left: posterior GP means (black) and confidence ellipses (blue) from data (red). Right: posterior given  rotated data. One can see that means and confidence ellipses rotate with the data.}
\end{figure*}

\section{Equivariant Stochastic Process Models}
\label{sec:equivariant_stochastic_process_model}
In this work, we are interested in  learning not only a single feature field $F$ but a probability distribution  $P$ over $\mathcal{F}_\rho$, i.e. a stochastic process over feature fields $F$. For example, $P$ could describe the distribution of all wind maps over a specific region. If $F\sim P$ is a random feature field and $g\in G$, we can define the transformed stochastic process $g.P$ as the distribution of $g.F$. We say that $P$ is \emph{$G$-invariant} if 
\begin{align}
 P=g.P \tforall g\in G
\end{align}
From a sample $F\sim P$, our model observes only a finite set of input-output pairs $Z=\{(\mathbf{x}_i,\mathbf{y}_i)\}_{i=1}^{n}$ where $\mathbf{y}_i$ equals $F(\mathbf{x}_i)$ plus potentially some noise. The induced representation naturally translates to a transformation of $Z$ under $G$ via
\begin{align}
 	\label{eq:trans_data_sets}
	g.Z :=\{(g\mathbf{x}_i,\rho(h)\mathbf{y}_i)\}_{i=1}^{n}
\end{align}

In a Bayesian approach, we can consider $P$ as a prior and given an observed data set $Z$ we can consider the posterior, i.e. the conditional distribution $P_{Z}$ of $F$ given $Z$. As the next proposition shows, equivariance of the posterior is the other side of the coin to invariance of the prior.
\begin{restatable}{proposition}{invprior}
\label{prop:inv_dist_leads_to_equiv}
Let $P$ be a stochastic process over $\mathcal{F}_\rho$. Then $P$ is $G$-invariant if and only if the  posterior map $Z\mapsto P_Z$ is $G$-equivariant, i.e. 
\begin{align}
    P_{g.Z}=g.P_{Z} \quad\text{for all } g\in G
\end{align}
\end{restatable}
The proof of this can be found in \cref{proof:invprior}.

In most real-world scenarios, it may not be possible to exactly compute the posterior and our goal is to build a model $Q$ which returns an approximation $Q_Z$ of $P_Z$. 
However, often it is our prior belief that the distribution $P$ is $G$-invariant. Given \cref{prop:inv_dist_leads_to_equiv}, it is then natural to construct an approximate inference model $Q$ which is itself equivariant. 

We will see applications of these ideas to GPs and CNPs in \cref{sec:equiv_GPs,sec:equiv_CNPs}.

\section{Equivariant Gaussian Processes}


\label{sec:equiv_GPs}
A widely-studied example of stochastic processes are Gaussian processes (GPs). Here we will look at Gaussian processes under the lens of equivariance. Since we are interested in vector-valued functions $F:\Reals^n\to\Reals^d$, we use matrix-valued positive definite kernels $K:\Reals^n\times\Reals^n\to\Reals^{d\times d}$ \citep{Review_matrix_kernels}.

In the case of GPs, we assume that for every $\mathbf{x},\mathbf{x}'\in\Reals^n$, it holds that $F(\mathbf{x})$ is normally distributed with mean $\mathbf{m}(\mathbf{x})$ and covariances $\text{Cov}(F(\mathbf{x}),F(\mathbf{x}'))=K(\mathbf{x},\mathbf{x}')$. We write $\mathcal{GP}(\mathbf{m},K)$ for the stochastic process defined by this.

We can fully characterise all mean functions and kernels leading to equivariant GPs:
\begin{restatable}{theorem}{equivgp}
\label{thm:equivgp}
A Gaussian process $\mathcal{GP}(\mathbf{m},K)$ is $G$-invariant, equivalently the posterior $G$-equivariant, if and only if
\begin{enumerate}
\item $\mathbf{m}(\mathbf{x})=\mathbf{m}\in\Reals^d$ is constant with $\mathbf{m}$ such that
\begin{align}
    \label{eq:m_invariant_under_rho}
    \rho(h)\mathbf{m}=\mathbf{m}\quad\text{for all } h\in H
\end{align}
\item $K$ fulfils the following two conditions:
\begin{enumerate}
\item $K$ is \textbf{stationary}, i.e. for all $\mathbf{x},\mathbf{x}'\in\Reals^n$
\begin{align}
\label{eq:GP_stationary}
K(\mathbf{x},\mathbf{x}')=K(\mathbf{x}-\mathbf{x}', \mathbf{0}) =: \hat{K}(\x - \x')
\end{align}
\item $K$ satisfies the \textbf{angular constraint}, i.e. for all $\mathbf{x},\mathbf{x}'\in\Reals^n,h\in H$ it holds that
\begin{align}
\label{eq:GP_kernel_constraint}
K(h \x, h \x') &= \rho(h) K(\x, \x')\rho(h)^T 
\end{align}
or equivalently, for all  $\mathbf{x}\in\Reals^n,h\in H$
\begin{align}
\hat{K}(h\mathbf{x})&=\rho(h)\hat{K}(\mathbf{x})\rho(h)^T
\end{align}
\end{enumerate}
If this is the case, we call $K$ $\rho$-equivariant.
\end{enumerate}
\end{restatable}
The proof of this can be found in \cref{proof:equivgp}.

We note the distinct similarity between the kernel conditions in \cref{eq:GP_stationary,eq:GP_kernel_constraint}, and the ones found in the equivariant convolutional neural network literature (see for example eq. (2) in \citet{E2Steerable}). In contrast to convolutional kernels, we have the additional constraint that the kernel $K$ must be positive definite. 

A popular example to model vector-valued functions is to simply use  $d$ independent GPs with a stationary scalar kernel $k:\Reals^n\to\Reals$. This leads to a kernel $K(\mathbf{x})=k(\mathbf{x})I$ which can easily be seen to be $E(n)$-equivariant.

As a non-trivial example of equivariant kernels, we will also consider the divergence-free and curl-free kernels (see \cref{sec:divcurl}) used in physics introduced by \citet{macedo2010learning} which allow us to model divergence-free and curl-free fields such as electric or magnetic fields (see      \cref{fig:gp_datasets}
 for examples).
 
We note that the kernels considered in  this work represent a small set of possible kernels permitted by \cref{thm:equivgp}. Much of the standard Gaussian Process and kernel machinery, e.g. Bochner's theorem, random Fourier features \citep{brault_random_2016}, and sparse methods, extend naturally to the vector-valued and equivariant case.

\section{Steerable Conditional Neural Processes}
\label{sec:equiv_CNPs}

Conditional Neural Processes were introduced as an alternative model to Gaussian processes. While GPs require us to explicitly model the prior $P$ and can perform exact posterior inference, CNPs aim to learn an approximation to the posterior map ($Z\mapsto P_Z$) directly, only implicitly learning a prior from data.
Generally speaking, the underlying architecture is a model which returns a mean function $\mathbf{m}_{Z}:\Reals^n\to\Reals^d$ and a covariance function $\mathbf{\Sigma}_{Z}:\Reals^n\to\Reals^{d\times d}$ given a context set $Z$. 
For simplicity, it makes the assumption that given $Z$ the functions values $F(\mathbf{x})$ are conditionally independent and normally distributed
\begin{align}
	\label{eq:CNP_model}
	F(\mathbf{x})\sim\mathcal{N}(\mathbf{m}_{Z}(\mathbf{x}),\mathbf{\Sigma}_{Z}(\mathbf{x})),\quad F(\mathbf{x})\indep F(\mathbf{x}') \; | \; Z
\end{align}

Let us call a model as in \cref{eq:CNP_model} a \emph{conditional process model}.
As we did with $F\in\mathcal{F}_\rho$, we can specify a law of transformation by considering $\mathbf{m}_Z$ as a \emph{mean feature field} in $\mathcal{F}_{\rho_m}$ and $\mathbf{\Sigma}_Z$ as a \emph{covariance feature field} in $\mathcal{F}_{\rho_{\Sigma}}$ for appropriate fiber representations $\rho_m,\rho_\Sigma$. With this, we can easily characterise equivariance in conditional process models:
\begin{restatable}{proposition}{SteerCNP}
\label{prop:SteerCNP}
A conditional process model  is $G$-equivariant if and only if the mean and covariance feature maps are $G$-equivariant, i.e. it holds for all $g\in G$ and context sets $Z$
\begin{align}
	\label{eq:mean_equiv_constraint}
	\mathbf{m}_{g.Z}&=g.\mathbf{m}_{Z}\\
		\label{eq:sigma_equiv_constraint}
	\mathbf{\Sigma}_{g.Z}&=g.\mathbf{\Sigma}_{Z}
\end{align}
with $\rho_{m}=\rho$ and $\rho_\Sigma=\rho\otimes\rho$ the tensor product with action given by
\begin{align}
    \label{eq:tensor_product_representation_sigma}
	\rho_{\Sigma}(h)A=\rho(h)A\rho(h)^T, \quad A \in \Reals^{d\times d}
\end{align} 
\end{restatable}

The proof can be found in \cref{proof:SteerCNP}.

In the following, we will restrict ourselves to perform inference from data sets of multiplicity $1$, i.e., data sets $Z=\{(\mathbf{x}_i,\mathbf{y}_i)\}_{i=1}^{m}$ where $\mathbf{x}_i \neq \mathbf{x}_j$ for all $i\neq j$. We denote the collection of all such data sets with $\mathcal{Z}_{\rho}$ meaning that they transform under $\rho$ (see 	\cref{eq:trans_data_sets}).

Moreover, we assume that there is no order in a data set $Z$, i.e. we aim to build models which are not only $G$-equivariant but also invariant to permutations of $Z$.
\begin{figure*}[t]%
    \centering
    \includegraphics[width=\textwidth]{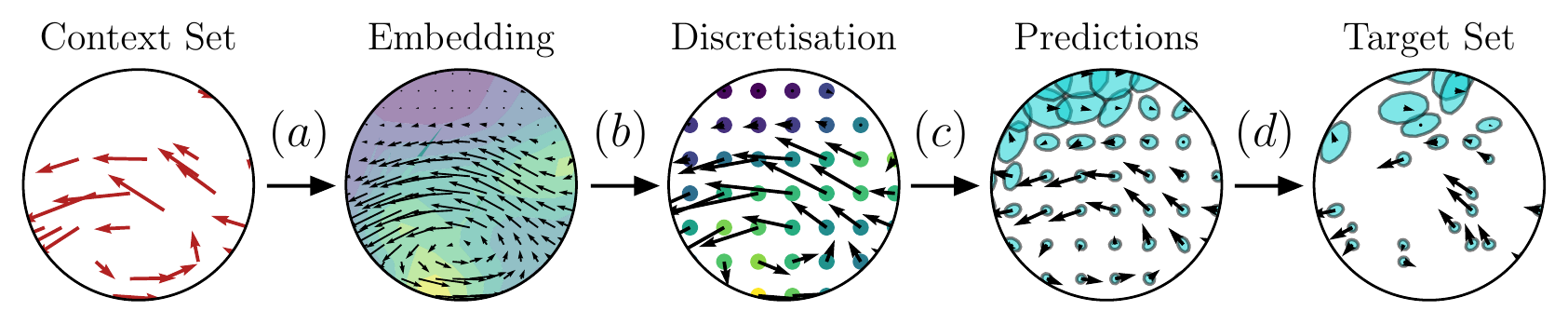}
    \caption{SteerCNP model illustration. (a) Embed the context set into a function. (b) Discretise this embedding on a regular grid. (c) Predict the mean and covariance of the conditional stochastic process on the grid of points. (d) Use kernel smoothing to predict the mean and covariance at target locations.}%
    \label{fig:feature_flow_SteerCNP}%
\end{figure*}
The following generalisation of the ConvDeepSets theorem of \citet{ConvCNP} gives us a universal form of all such conditional process models. We simple need to pick $\rho_{\tin} = \rho$ and $\rho_{\tout} = \rho_{m} \oplus \rho_{\Sigma}=\rho\oplus \rho_{\Sigma}$ in the following theorem.
\begin{restatable}[EquivDeepSets]{theorem}{equivdeepsets}
\label{thm:equivdeepsets}
Let $\rho_\tin,\rho_\tout$ be the two fiber representations. Define the embedding representation as the direct sum $\rho_E = \rho_{triv} \oplus \rho_{\tin}$.\\
A function ${\Phi}: \ZZ_{\rho_{\tin}} \to\mathcal{F}_{\rho_{\tout}}$ is $G$-equivariant and permutation invariant if and only if it can be expressed as
\begin{align}
    {\Phi}(Z) = \Psi(E(Z))
\end{align}
for all $Z=\{(\mathbf{x}_i,\mathbf{y}_i)\}_{i=1}^{m}\in\mathcal{Z}_{\rho_{\tin}}$ with
\begin{enumerate}
\item     $E(Z) = \sum_{i=1}^m  K(\cdot, \x_i)\phi(\y_i)$ 
\item $\phi(\mathbf{y})=(1,\mathbf{y})^T\in\Reals^{d+1}$.
\item $K: \Reals^n \times \Reals^n \to \Reals^{(d+1) \times(d+1)}$ is a $\rho_E$-equivariant strictly positive definite kernel (see \cref{thm:equivgp}).
\item $\Psi:  \mathcal{F}_{\rho_E} \to \mathcal{F}_{\rho_{\tout}}$ is a $\grp$-equivariant function. 
\end{enumerate}
Additionally, by imposing extra constraints (see \cref{proof:equivdeepsets}), we can also ensure that $\Phi$ is continuous.
\end{restatable}

The proof of this can be found in appendix \ref{proof:equivdeepsets}. Using this, we can start to build SteerCNPs by building an encoder $E$ and a decoder ${\Psi}$ as specified in the theorem.

The form of the encoder only depends on the choice of a kernel $K$ which is equivariant under $\rho_E$. An easy but effective way of doing this is to pick a kernel $K_0$ which is equivariant under $\rho$ (see \cref{sec:equiv_GPs}) and a scalar kernel $k:\Reals^n\to\Reals$ and then use the block-version $K=k\oplus K_0$.

\subsection{Decoder}
\label{sec:decoder}
By \cref{thm:equivdeepsets}, it remains to construct a $G$-equivariant decoder $\Psi$. To construct such maps, we will use steerable CNNs \citep{SteerCNNs,E2Steerable,3DSteerable}. In theory, a layer of such a network is an equivariant function $\Psi:\mathcal{F}_{\rho_\tin}\to\mathcal{F}_{\rho_\tout}$ where we are free to choose fiber representations $\rho_\tin,\rho_\tout$.

Steerable convolutional layers are defined by a kernel $\kappa:\Reals^n\to\Reals^{c_\tout\times c_\tin}$ such that the map
\begin{align}
	\label{eq:equiv_convolution}
	[\kappa\star F](\mathbf{x})=\int\kappa(\mathbf{x}, \mathbf{x}')F(\mathbf{x}')d\mathbf{x}
\end{align} 
is $G$-equivariant. These layers serve as the learnable, parameterisable functions.

Steerable activation functions are applied pointwise to $F$. These are functions $\sigma:\Reals^{c_\tin}\to\Reals^{c_\tout}$ such that
\begin{align}
	\label{eq:equiv_activation_function}
	\sigma(\rho_{\tin}(h)\mathbf{x})=\rho_{\tout}(h)\sigma(\mathbf{x})
\end{align}
As a decoder of our model, we use a stack of equivariant convolutional layers composed with equivariant activation functions. The convolutions in \cref{eq:equiv_convolution} are computed in a discretised manner after sampling $E(Z)$ on a grid $\mathcal{G}\subset\Reals^n$. Therefore, the output of the neural network will be a discretised version of a function and we use kernel smoothing to extend the output of the network to the whole space.
\subsection{Covariance Activation Functions}
The output of a steerable neural network has general vectors in $\Reals^c$ for some $c$ as outputs. Therefore, we need an additional component to obtain (positive definite) covariance matrices in an equivariant way.

We introduce the following concept:
\begin{definition}
An \textbf{equivariant covariance activation function} is a map $\eta:\Reals^{c}\to\Reals^{d\times d}$ together with a fiber representation $\rho_{\eta}:H\to\GL(\Reals^{c})$ such that for all $\mathbf{y}\in\Reals^c$ and $h\in H$
\begin{enumerate}
 \item $\eta(\mathbf{y})$ is a symmetric, positive semi-definite matrix.
\item $\eta(\rho_\eta(h)\mathbf{y})=\rho_{\Sigma}(h)\eta(\mathbf{y})$
\end{enumerate}
\end{definition}
In our case, we use a \emph{quadratic covariance activation function} which we define by
\begin{align*}
	\eta:\Reals^{d\times d}\to\Reals^{d\times d},\quad \eta(A)=AA^T
\end{align*}
Considering $A=(a_1,\dots,a_D)\in\Reals^{d^2}$ as a vector by stacking the columns, the input representation is then $\rho_{\eta}=\rho\oplus\dots\oplus\rho$ as the $d$-times sum of $\rho$. It is straight forward to see that $\eta$ is equivariant and outputs positive semi-definite matrices.
\subsection{Full model}
Finally, we summarise the architecture of the SteerCNP (see \cref{fig:feature_flow_SteerCNP}):
\begin{enumerate}
\item The \textbf{encoder} produces an embedding of a data set $Z$ as a function $E(Z)$.
\item A discretisation of $E(Z)$ serves as input for the \textbf{decoder}, a steerable CNN with input fiber representation $\rho_{E}$ and output fiber representation $\rho\oplus\rho_{\eta}$.
\item On the covariance part, we apply the covariance activation function $\eta$.
\item The grid values of the mean and the covariances are extended to the whole space $\Reals^n$ via kernel smoothing.
\end{enumerate}
We train the model similar to the CNP by iteratively sampling a data set $Z$ and splitting it randomly in a context set $Z_C$ and a target set $Z_T$. The context set $Z_C$ is then passed forward through the SteerCNP model and the mean log-likelihood of the target $Z_T=\{(\mathbf{x}'_i,\mathbf{y}'_i)\}_{i=1}^{m}$ is computed. In brief, we minimise the loss
\begin{align*}
	-\mathbb{E}_{Z_C,Z_T\sim P}\left[\frac{1}{m}
		\sum\limits_{i=1}^{m}\log\mathcal{N}(\mathbf{y}_i';\mathbf{m}_{Z_C}(\mathbf{x}'_i),\mathbf{\Sigma}_{Z_C}(\mathbf{x}'_i))
	\right]
\end{align*}
by gradient descent methods.

In sum, this gives a CNP model, which up to discretisation errors is equivariant with respect to arbitrary transformations from the group $G$ and invariant to permutations.

\section{Related Work} 
\label{sec:related_work}
\textbf{Equivariance and symmetries in deep learning.}
Motivated by the success of the translation-equivariant CNNs \citep{Classical_CNNs}, there has been a great interest in building neural networks which are equivariant also to more general transformations. Approaches use a wide range of techniques such as convolutions on groups \citep{SphericalCNNs,KondTrivGen,G_CNNs,HexaConv,CubeNet}, cylic permutations \citep{CyclicSymmetry}, Lie groups \citep{LieGroup} or phase changes \citep{HarmonicNetworks}. 
It was in the context of Steerable CNNs and its various generalisations where ideas from physics about fields started to play a more prominent role. \citep{SteerCNNs,3DSteerable,E2Steerable,Homogenous_spaces}. We use this framework since it allows for modelling of non-trivial transformations of features $F(\mathbf{x})$ via fiber representations $\rho$, as for example necessary to model transformations of vector fields (see \cref{eq:vf_transformation}). 

\textbf{Gaussian Processes and Kernels.} Classical GPs using kernels such as the RBF or Mat\'ern kernels have been widely used in machine learning to model scalar fields \citep{GP_book}. Vector-valued GPs, which allow for dependencies across dimensions via matrix-valued kernels, were common models in geostatistics \citep{geostatistics} and also played a role in kernel methods  \citep{Review_matrix_kernels}. With this work, we showed that many of these GPs are equivariant (see \cref{thm:equivgp}) giving a further theoretical foundation for their applicability. In contrast to equivariant GPs, \citet{Reissert} consider the construction of equivariant functions from kernels, arriving at similar equivariance constraints as we do in \cref{thm:equivgp}.

\textbf{Neural Processes.} \citet{CNPs} introduced Conditional Neural Processes (CNPs) as an architecture constructed out of neural networks which learns an approximation of stochastic processes from data. They share the motivation of meta-learning methods \citep{MAML,GradDescentbyGradDescent} to learn a distribution of tasks instead of only a single task. Neural Processes (NPs) are the latent variable counterpart of CNPs allowing for correlations across the marginals of the posterior. Both CNPs and NPs have been combined with other machine learning concepts, for example attention mechanisms \citep{attNP}. 

\citet{ConvCNP} were the first to consider symmetries in NPs. Inspired by the prevalence of stationary kernels in the GP literature, they introduced a translation-equivariant NP model, along with a universal characterisation of such models. Our work can be seen as a generalisation of their method. By picking a trivial fiber group $H=\{e\}$ and $K$ as a diagonal RBF-kernel in SteerCNPs (see  \cref{thm:equivdeepsets}), we get the ConvCNP as a special case of SteerCNPs. 

During the development of this work, \citet{EquivCNPs} also studied more general equivariance in CNPs, restricting themselves to scalar fields and focusing on Lie Groups. They only compared their approach with previous models on $1d$ synthetic regression tasks where their approach did not seem to outperform ConvCNPs, and in this case the added symmetry is redundant. In contrast, we focus on general scalar and vector-valued fields with non-trivial transformations $\rho$. As we show in the next section, this "steerable" approach leads to significant performance gains compared to previous models. The decomposition theorem presented in \citet{EquivCNPs} can be seen as a special case of  \cref{thm:equivdeepsets} in this work by setting $\rho = \rho_{tri}$.

\section{Experiments}

\label{sec:experiments}
Finally, we provide empirical evidence that equivariance is a helpful bias in stochastic process models. We focus on evaluating SteerCNPs since inference with the variety of GPs, which this work shows to be equivariant, has been studied exhaustively. 

\subsection{Gaussian Process Vector Fields}

A common baseline task for CNPs is regression on samples from a Gaussian process \citep{CNPs,ConvCNP}, partially because one can directly compare the output of the model with the true posterior. Here, we consider the task of learning $2$D vector fields $F:\Reals^2\to\Reals^2$ which are samples of a Gaussian process $\mathcal{GP}(0,K)$ with 3 different $E(2)$-equivariant kernels $K$:  the diagonal RBF-kernel, the divergence-free kernel and the curl-free kernel (see \cref{fig:gp_datasets,sec:divcurl}).

We run extensive experiments comparing the SteerCNP with the CNP and the translation-equivariant ConvCNP. On the SteerCNP, we impose various levels of rotation and reflection equivariance by picking different fiber groups $H$. As usual for CNPs, we use the mean log-likelihood as a measure of performance. The maximum possible log-likelihood is obtained by Monte Carlo sampling using the true GP posterior.

In \cref{table:GP_results}, the results are presented. Overall, one can see that the SteerCNP clearly outperforms previous models by reducing the difference to the GP baseline by more than a half. In addition, we observe that small fiber groups ($H=C_4$) lead to the best results. Although theoretically models with the largest fiber groups ($H=C_{16},SO(2)$) should perform better, it is possible that practical limitations such as discretisation of the model favors smaller fiber groups since they still allow for compensation of marginal asymmetries and numerical errors. For the case of $H=SO(2)$, the worse results are consistent with results from \citet{E2Steerable} in supervised learning and practical reasons for this are discussed in more depth there. Hence, we leave out $SO(2)$ in further experiments.

\begin{table}[t]
    \small
    \centering
    \setlength\tabcolsep{3pt}
    \caption{Results for experiments on GP vector fields. Mean log-likelihood $\pm$ 1 standard deviation over 5 random seeds reported. Last row shows GP-baseline (optimum).}
    \vspace{0.1in}
    \begin{tabular}[b]{lrrr}
        \toprule
        \textbf{Model} & RBF & Curl-free & Div-free \\
        \midrule
        CNP    & -4.24$\pm$0.00& -0.750$\pm$0.004& -0.752$\pm$0.006 \\
        ConvCNP &-3.88$\pm$0.01& -0.541$\pm$0.004& -0.533$\pm$0.001\\
        \midrule
        SteerCNP ($SO(2)$)   & -3.93$\pm$0.05& -0.550$\pm$0.005& -0.552$\pm$0.008 \\ 
        SteerCNP ($C_4$)       & \textbf{-3.66}$\pm$0.00& \textbf{-0.461}$\pm$0.003 &\textbf{-0.464}$\pm$0.007\\
        SteerCNP ($C_8$)       & -3.70$\pm$0.01& -0.479$\pm$0.004& -0.478$\pm$0.004 \\
        SteerCNP ($C_{16}$)    & -3.71$\pm$0.02& -0.476$\pm$0.005&-0.480$\pm$0.008\\
        SteerCNP ($D_4$)       & -3.72$\pm$0.03& -0.471$\pm$0.002& -0.477$\pm$0.005 \\
        SteerCNP ($D_8$)     & \textbf{-3.68}$\pm$0.03& \textbf{-0.462}$\pm$0.005& -0.467$\pm$0.008\\
        \midrule
        GP & -3.50 &-0.410 & -0.411\\
        \bottomrule
    \end{tabular}
    \label{table:GP_results}
\end{table}

\subsection{Image in-painting}

To test the SteerCNP on purely scalar data we evaluate our model on an image completion task. In image completion tasks the context set is made up of pairs of 2D pixel locations and pixel intensities and the objective is to predict the intensity at new locations. For further details see \cref{sec:image_details}, along with qualitative results from the experiments.

\textbf{MNIST and rotMNIST.}
We first train models on completion tasks from the MNIST data set \citep{lecun2010mnist}. The results in \cref{tab:mnist} show that the equivariance built into the SteerCNP is useful, with the various SterableCNPs outperforming previous models. To test whether this performance gain could be replicated by data augmentation, we trained the models on a variant, called rotMNIST, produced by randomly rotating each digit in the dataset. We find that this does not improve results in the non equivariant models, in fact causing a decrease in performance as they try to learn a more complex distribution from too little data. It therefore seems to be the gain in parameter and data efficiency due the imposed equivariance constraints what leads to better performance of SteerCNPs.

\textbf{Generalising to larger images.}
One key advantage of equivariant models is that they should be able to extrapolate to data at locations and orientations previously unseen. To test this, we create a new data set by sampling 2 rotMNIST images and randomly translating them in a 56*56 image, calling this dataset extMNIST (for extrapolate). We evaluate the models trained on MNIST on this data set. The results in \cref{tab:mnist} show the significant benefit of the additional rotation equivariance.\footnote{We also conducted experiments with other train-test combinations of rotMNIST (resp. extMNIST). The results confirm our observations and can be found in \cref{sec:image_details}.}

\begin{figure}[t]
    \centering
    \includegraphics[width=0.75\linewidth]{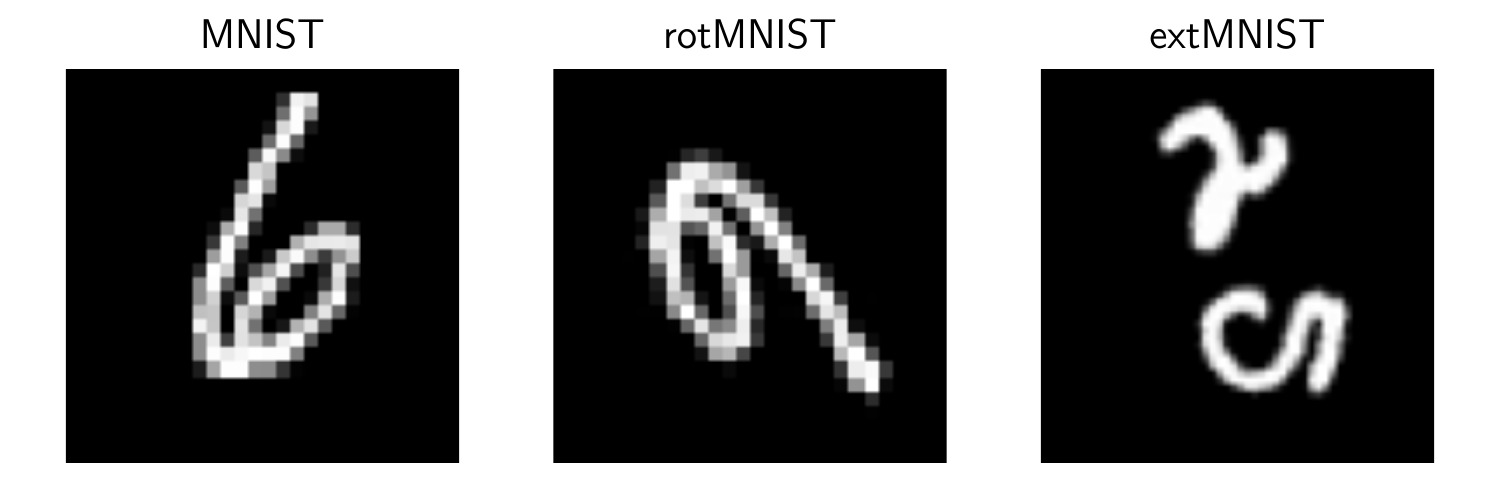}
    \caption{Examples of the three datasets}
\end{figure}
\begin{table}[t]
\small
\centering
\caption{Results for the MNIST experiments. Mean log-likelihood $\pm$ 1 standard deviation over 3 random model and dataset seeds reported.}
\vspace{0.1in}
\setlength\tabcolsep{3pt}
\begin{tabular}{lrrr}
\toprule
\textbf{Train dataset} &                   \multicolumn{1}{l}{MNIST} &                \multicolumn{1}{l}{rotMNIST} &                   \multicolumn{1}{r}{MNIST} \\
\textbf{Test dataset} & \multicolumn{1}{l}{MNIST} &\multicolumn{1}{l}{MNIST} &  {extMNIST} \\
\textbf{Model}            &                         &                         &                      \\
\midrule
GP                 &         {0.39$\pm$0.30} &         {0.39$\pm$0.30} &         {0.72$\pm$0.17} \\
CNP                &         {0.76$\pm$0.05} &         {0.66$\pm$0.06} &        {-1.11$\pm$0.06} \\
ConvCNP            &         {1.01$\pm$0.01} &         {0.95$\pm$0.01} &         {1.08$\pm$0.02} \\
\midrule
SteerCNP($C_4$)    &         {1.05$\pm$0.02} &  \textbf{1.02$\pm$0.03} &         {1.14$\pm$0.02} \\
SteerCNP($C_{8}$)  &  \textbf{1.07$\pm$0.03} &  \textbf{1.05$\pm$0.04} &  \textbf{1.16$\pm$0.03} \\
SteerCNP($C_{16}$) &  \textbf{1.08$\pm$0.03} &  \textbf{1.04$\pm$0.03} &  \textbf{1.17$\pm$0.05} \\
SteerCNP($D_4$)    &  \textbf{1.08$\pm$0.03} &  \textbf{1.05$\pm$0.03} &         {1.14$\pm$0.03} \\
SteerCNP($D_8$)    &  \textbf{1.08$\pm$0.03} &  \textbf{1.04$\pm$0.04} &  \textbf{1.17$\pm$0.02} \\
\bottomrule
\end{tabular}
\label{tab:mnist}
\end{table}

\subsection{ERA5 Weather Data}

\begin{figure*}[h]
    \centering
    \includegraphics[width=0.9\textwidth]{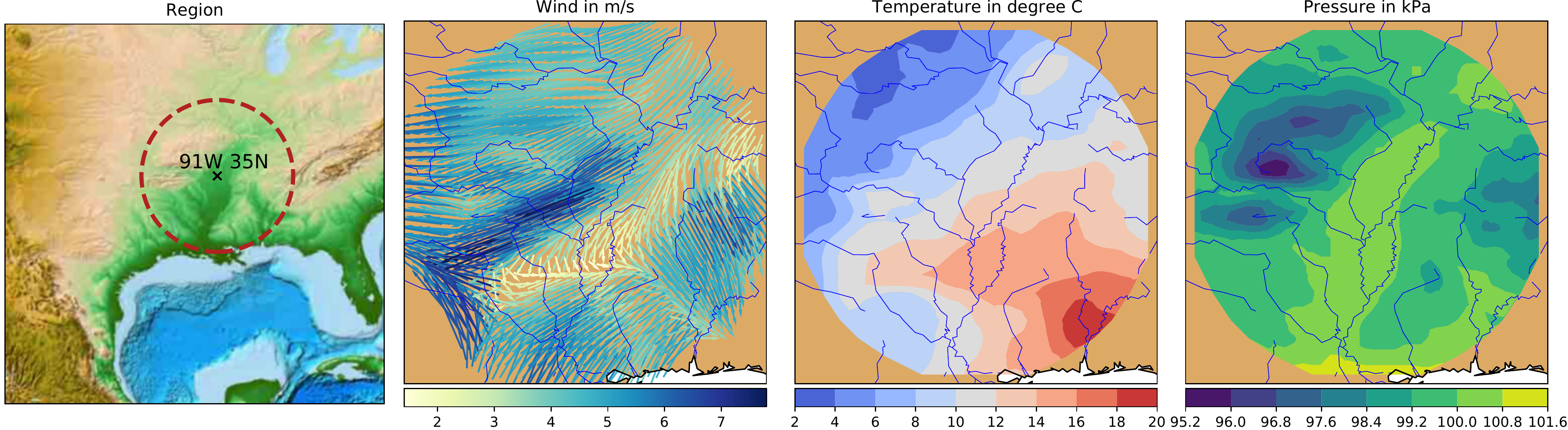}
  \caption{Illustration of ERA5 weather data from the US.}
    \label{fig:ERA5_US_Illustration}
\end{figure*}

To evaluate the performance of the SteerCNP model on vector-valued fields, we retrieved weather data from the global ERA5 data set.\footnote{We obtained this data using Copernicus Climate Change Service Information [2020].} We extracted data from a circular region surrounding Memphis, Tennessee, and from a region of the same size in Hubei province, Southern China (see \cref{fig:ERA5_US_Illustration} for illustration and \cref{sec:Experimental details} for details). While the specific choice of these areas was arbitrary, we tried to pick two topologically different regions far away from each other.
 
Every sample $F$ from these data sets corresponds to a weather map consisting of temperature, pressure and wind in the region at one single point in time. 
We give the models the task to infer a wind vector field  from a data $Z$ of pairs $(\mathbf{x},\mathbf{y})$ where $\mathbf{y}=(y^t,y^p,y^w_1,y^w_2)\in\Reals^4$ gives the temperature, pressure and wind at point $\mathbf{x}$. In particular, the output features are only a subset of the input features. To deal with such a task, we can simply pick different input and output fiber representations for the SteerCNP:
\begin{align*}
	\rho_\tin=\rho_{triv}\oplus\rho_{triv}\oplus\rho_{\Id},\quad \rho_\tout=\rho_{\Id}
\end{align*}
\textbf{Performance on US data.} As a first experiment, we split the US data set in a train, validation and test data set. Then we train and test the models accordingly. We observe that the SteerCNP outperforms previous models like a GP with RBF-kernel, the CNP and the ConvCNP with a significant margin for all considered fiber groups (see \cref{table:ERA5_results}). Again, we observe that a relatively small fiber group $C_4$ leads to the best results. Inference from weather data is clearly not exactly equivariant due to local differences such as altitude and distance to the sea. Therefore, it seems that a SteerCNP model with small fiber groups like $C_4$ enables us to exploit the equivariant patterns much better than the ConvCNP and CNP but leaves flexibility to account for asymmetric patterns.

\textbf{Generalising to a different region.}
As a second experiment, we take these models and test their performance on data from China. This can be seen as a transfer learning task. Intuitively, posing a higher equivariance restriction on the model makes it less adapting to special local circumstances and more robust when transferring to a new environment. Indeed, we observe that the CNP, the ConvCNP and the model with fiber group $C_4$ have a larger loss in performance than SteerCNP models with larger fiber groups such as $C_{16},D_{8},D_4$. Similarly, while GPs had a significantly worse performance than ConvCNPs on the US data, it outperforms it on the transfer to China data. In applications like robotics where environments constantly change this robustness due to equivariance might be advantageous.
\begin{table}[t]
    \setlength\tabcolsep{3pt}
    \centering
    \small
    \caption{Results on ERA5 weather experiment trained on US data. Mean log-likelihood $\pm$ 1 standard deviation over 5 random seeds reported. Left: tested on US data. Right: tested on China data.}
    \vspace{0.1in}
    \begin{tabular}[b]{lrr}
        \toprule
        \textbf{Model} & US & China  \\
        \midrule
        GP & 0.386$\pm$0.005& -0.755$\pm$0.001\\
        CNP    & 0.001$\pm$0.017& -2.456$\pm$0.365 \\
        ConvCNP & 0.898$\pm$0.045& -0.890$\pm$0.059\\
        \midrule
        SteerCNP ($C_4$)       & \textbf{1.255}$\pm$0.019& -0.578$\pm$0.173\\
        SteerCNP ($C_{8}$)    & 1.038$\pm$0.026& -0.582$\pm$0.104\\
        SteerCNP ($C_{16}$)    & 1.094$\pm$0.015& -0.550$\pm$0.073\\
        SteerCNP ($D_4$)       & 1.037$\pm$0.037& \textbf{-0.429}$\pm$0.067 \\
        SteerCNP ($D_8$)     & 1.032$\pm$0.011& -0.539$\pm$0.129\\
        \bottomrule
    \end{tabular}
    \label{table:ERA5_results}
\end{table}

\section{Limitations and Future Work}
Similar to CNPs, our model cannot capture dependencies between the marginals of the posterior. Recently, \citet{ConvNP} introduced a translation-equivariant NP model which allows to do this and future work could combine their approach with general symmetries considered in this work.


As stated earlier, our model works for Euclidean spaces of any dimension. The limiting factor is the development of Steerable CNNs used in the decoder. In our experiments, we focused on $\RR^2$ as the code is well developed by \citet{E2Steerable}. Recent developments in the design of equivariant neural networks also explore non-Euclidean spaces such as spheres and encourage exploration in this direction in the context of NPs \citep{Homogenous_spaces,SphericalCNNs,SphericalVFs}. Additionally, in Euclidean space we can incorporate other symmetries such as scaling via Lie group approaches, similar to \citep{EquivCNPs}, or symmetry to uniform motion  for fluid flow \citep{wang2020incorporating} by choosing suitable equivariant CNNs.

One practical limitation of this method is the necessity to discretise the continuous RKHS embedding, which can be costly and breaks the theoretical guarentees found in this work in \cref{thm:equivdeepsets}. An alternative approach would be to move away entirely from the structure of \cref{thm:equivdeepsets}, and build an architecture more similar to the original CNP, utilising equivariant point cloud methods \citep{finzi2020generalizing, hutchinson2020lietransformer, satorras2021n} to produce the embedding of the context set. This would avoid the discretisation of the embedding grid and may provide speedups, but looses the flexibility promised by \cref{thm:equivdeepsets}.


\section{Conclusion}
\label{sec:conclusion}
In this work, we considered the problem of learning stochastic fields and focused on using their geometric structure. We motivated the design of equivariant stochastic process models by showing the equivalence of equivariance in the posterior map to invariance in the prior data distribution. We fully characterised equivariant Gaussian processes and introduced Steerable Conditional Neural Processes, a model that combines recent developments in the design of equivariant neural networks with the family of Neural Processes. We showed that it improves results of previous models, even for data which shows inhomogeneities in space such as weather, and is more robust to perturbations in the underlying distribution.

Our work shows that implementing general symmetries in stochastic process or meta-learning models could be a substantial step towards more data-efficient and adaptable machine learning models. Inference models which respect the structure of fields, in particular vector fields, could further improve the application of machine learning in natural sciences, engineering and beyond.


\subsection*{Acknowledgements}

Peter Holderrieth is supported as a Rhodes Scholar by the Rhodes Trust. Michael Hutchinson is supported by the EPSRC Centre for Doctoral Training in Modern Statistics and Statistical Machine Learning (EP/S023151/1). Yee Whye Teh’s research leading to these results has received funding from the European Research Council under the European Union’s Seventh Framework Programme (FP7/2007-2013) ERC grant agreement no. 617071.

We would also like to thank
the Python community~\citep{van1995python,oliphant2007python} for developing
the tools that enabled this work, including Pytorch \citep{paszke2017automatic},
{NumPy}~\citep{oliphant2006guide,walt2011numpy, harris2020array},
{SciPy}~\citep{jones2001scipy}, and
{Matplotlib}~\citep{hunter2007matplotlib}.

\bibliography{SteerCNPs}
\bibliographystyle{icml2021}

\clearpage
\appendix
\onecolumn
\section{Basics for Group and Representation Theory}

This section gives the basic definitions about groups and representations necessary to understand this work. We refer to the literature for a more detailed introduction \citep{artin2011algebra,GroupRepRef}.
\label{sec:intro_to_groups}
\subsection{Groups}

A \textit{group} $(G,\op)$ is a set $G$ together with a function $\op:G\times G\to G, (g,h)\mapsto g\op h$ called group \emph{operation} satisfying
\begin{enumerate}
\item (Associativity): $g\op(h\op i)=(g\op h)\op i$ for all $g,h,i\in G$
\item (Existence of a neutral element): There is a $e\in G$ such that: $e\op g=g\op e= g$ for all $g\in G$
\item (Existence of an inverse): For all $g\in G$, there is a $g^{-1}$ such that $e=g^{-1}\op g=g\op g^{-1}$
\end{enumerate}
If in addition, $G$ satisfies
\begin{enumerate}[resume]
	\item (Commutativity): $g\op h=h\op g$ for all $g,h\in G$
\end{enumerate}
$G$ is called \emph{Abelian}. We simply write $g_1g_2$ for $g_1\op g_2$ if it is clear from the context.

If $\rho:G\to G'$ is a map between two groups, it is called a \emph{group homomorphism} if $\rho(g\op g')=\rho(g)\op\rho(g')$. That is, the map preserves the action of the group. A \emph{group isomorphism} is a homomorphism that is bijective. In the later case, $G$ and $G'$ are called \text{isomorphic} and we write $G\cong G'$. 

\textbf{The Euclidean group}

In the context of this work, the most important example of a group is the \emph{Euclidean group} $E(n)$ consisting of all \emph{isometries}, i.e. the set of all functions $T:\Reals^n\to\Reals^n$ such that 
\begin{align*}
    \|T(\mathbf{x})-T(\mathbf{x}')\|=\|\mathbf{x}-\mathbf{x}'\|,\tforall \mathbf{x},\mathbf{x}'\in\Reals^n
\end{align*}
Defining the group operation as the composition of two isometries by $T_1\op T_2:=T_1\circ T_2$, we can identify $E(n)$ as a group.

\textbf{Subgroups}

A \textit{subgroup} $H$ of a group $(G,\cdot)$ is a subset $H\subset G$ which is closed under the action of the original group. I.e. a set $\hrp\subset \grp$ is a subgroup of $(\grp, \cdot)$ if $h_1 \op h_2\in\hrp$ for all $h_1, h_1 \in \hrp$ and $h^{-1}\in H$ for all $h\in H$. A subgroup is typically denoted by $\hrp < \grp$.

 We can identify all intuitive geometric transformations on $\Reals^n$ as subgroups of $E(n)$:
\begin{enumerate}
\item \textbf{Translation:} For any vector $\mathbf{x}\in\Reals^n$, a translation by $\mathbf{x}$ is given by the map $t_\mathbf{x}:\Reals^n\to\Reals^n,\mathbf{x}'\mapsto \mathbf{x}+\mathbf{x}'$. The group of all translations is denoted by $T(n)$ .
\item \textbf{Rotoreflection:} The orthogonal group $O(n)=\{Q\in\Reals^{n\times n}| QQ^T=I\}$ describes all reflections and subsequent rotations.
\item \textbf{Rotation:} The special orthogonal group $SO(n)=\{R\in O(n)| \det R=1\}$ describes all rotations in $\Reals^n$. 
\end{enumerate}

\textbf{Normal subgroups}

A \textit{normal subgroup} of a group is a subgroup which is closed under conjugation of the group. That is, $N$ is a normal subgroup of $\grp$ if it is a subgroup of $\grp$ and
\begin{equation*}
    gng\inv \in N \; \tforall \; n \in N, g \in \grp
\end{equation*}
Typically a normal subgroup is denoted $N \triangleleft \grp$. The most important example for this work is $T(n)\triangleleft  E(n)$.

\newpage
\textbf{Semidirect product groups}
\hypertarget{def:semidicrectproduct}{}

A group $G$ is a \textit{semidirect product} of a subgroup $\hrp<\grp$ and a normal subgroup $N\triangleleft\grp$ if it holds that for all $g\in G$, there are unique $n\in N,h\in H$ such that $g=nh$.
There are a number of equivalent conditions, but not needed for this exposition. The semidirect product of two groups is denoted by
\begin{equation*}
    \grp = N \rtimes \hrp
\end{equation*}
Most importantly, we can identify $E(n)=T(n)\rtimes O(n)$ as the semidirect product of $T(n)$ and $O(n)$.
\subsection{Representations of Groups}
Group representations are a powerful tool to describe the algebraic properties of geometric transformations:
\hypertarget{def:representation}{}
Let $V$ be a vector space and $\GL(V)$ be the \textbf{general linear group}, i.e. the group of all linear, invertible transformations on $V$ with the composition $f\op g=f\circ g$ as group operation. Then a \textbf{representation} of a group $H$ is a group homomorphism $\rho:\hrp\to \GL(V)$.
For $V=\RR^d$, this is the same as saying a group representation is a map $\rho: \hrp \to \RR^{d\times d}$ such that 
\begin{align*}
    \rho(h_1\op h_2)=\rho(h_1) \rho(h_2) 
\end{align*}
where the right hand side is typical matrix multiplication.

The simplest group representation is the \hypertarget{trivialrep}{\textit{trivial representation}} $\rho_{triv}$ which maps all elements of the group to the identity,
\begin{align}
    \rho_{triv}(h) =\mathbf{1}_{d} \tforall h\in H
\end{align}

\textbf{Orthogonal and unitary groups} 

An \hypertarget{def:orthogonal_representation}{\textit{orthogonal representation}} is a representation $\rho:H\to\GL(\RR^d)$ such that $\rho(h)\in O(d)$ for all $h\in H$. For \textit{compact groups} $H$, every representation is equivalent to an orthogonal  representation \citep[Theorem II.1.7]{GroupRepRef}. This is useful as the identity $\rho(h)^T = \rho(h)\inv$ often makes calculations significantly easier. Since in this work we focus on subgroups $H\subset O(d)$ which are all compact, it is not a restriction to assume that.

\textbf{Direct sums} 

Given two representations, $\rho_1: \hrp \to \GL(\RR^n)$ and $\rho_2: \hrp \to \GL(\RR^m)$ , we can combine them together to give their \textit{direct sum}, $\rho_1 \oplus \rho_2: H\to \GL(\RR^{n+m})$, defined by
\begin{equation}
    (\rho_1 \oplus \rho_2)(h) = \begin{bmatrix} \rho_1(h) & 0 \\ 0 & \rho_2(h) \end{bmatrix}
\end{equation}
i.e the block diagonal matrix comprised of the two representations. This sum generalises to an arbitrary number of representations.

\hypertarget{def:tensor_product}{\textbf{Tensor products}} 

Let $V_1,V_2$ be two vector spaces and $V_1\otimes V_2$ their tensor product. Given two representations, $\rho_1: \hrp \to \GL(V_1)$ and $\rho_2: \hrp \to \GL(V_2)$, we can take the tensor product representation $\rho_1 \otimes \rho_2: \hrp \to \GL(V_1 \otimes V_2)$ defined by the condition that 
\begin{align}
    [\rho_1 \otimes \rho_2](h)(v_1 \otimes v_2) = (\rho_1(h)v_1) \otimes (\rho_2(h)v_2)
\end{align}
for all $v_1\in V_1$, $v_2\in V_2$, $h \in \hrp$.

To make this concrete for \cref{prop:SteerCNP}, we have $V_1,V_2=\RR^d$ and the tensor product becomes $V_1\otimes V_2=\RR^{d\times d}$ with $v_1\otimes v_2=v_1v_2^T$ the outer product for all $v_1,v_2\in\RR^d$. Setting $\rho=\rho_1=\rho_2$, the tensor product representation $\rho\otimes\rho$ becomes
\begin{align}
    [\rho\otimes\rho](h)(v_1\otimes v_2)=(\rho(h)v_1)(\rho(h)v_2)^T
    =\rho(h)v_1v_2^T\rho(h)^T
\end{align}
Therefore, $\rho_{\Sigma}$ as defined in \cref{eq:tensor_product_representation_sigma} is the tensor product $\rho\otimes\rho$.
We can return such representations to the more usual matrix-acting-on-vector format by vectorising these expressions. Using the identity $\vect(ABC) = [C^T \otimes_{kron} A] \vect(B)$, with $\otimes_{kron}$ being the usual Kronecker product and $\vect(A)$ being the column-wise vectorisation of $A$ we get 
\begin{align}
    \vect(\rho(h) A \rho(h)^T) = [\rho(h) \otimes_{kron} \rho(h)] \vect(A)
\end{align}

\section{Proofs}
\subsection{Proof of \cref{prop:inv_dist_leads_to_equiv}
}
\label{proof:invprior}
\invprior*
\begin{proof}
Let us be given a distribution $P$ over functions $\mathcal{F}_\rho$ and $F\sim P$. Define $g.P$ to be the distribution of $g.F$. For any $\mathbf{x}_1,\dots,\mathbf{x}_k\in\Reals^n$ let $\mathbf{x}_{1:k}$ denote the concatenation $(\mathbf{x}_1,\dots,\mathbf{x}_k)$ 
of these vectors and let $g\mathbf{x}_{1:k}$ be $(g\mathbf{x}_1,\dots,g\mathbf{x}_k)$.
For any such $\mathbf{x}_{1:k}$, let $\psi_{\mathbf{x}_{1:k}}^P$ be the finite-dimensional marginal of $P$, i.e. the distribution such that
\begin{align*}
    [F(\mathbf{x}_1),\dots,F(\mathbf{x}_k)]^T\sim \psi_{\mathbf{x}_{1:k}}^P
\end{align*}
For simplicity, we assume here that $\psi_{\mathbf{x}_{1:k}}^{P}$ is absolutely continuous with respect to the Lebesgue measure, i.e. has a density $\lambda_{\mathbf{x}_{1:k}}^P$. Our proof uses Kolmogorov's theorem \citep{KolmogorowReference}, which says that two stochastic processes coincide if and only if their finite-dimensional marginals agree. Before the actual proof, we need the following four auxiliary statements.

\textbf{1. Marginals of posterior.} Let $F\sim P$ and let us given a context set $Z=\{(\mathbf{x}_i',\mathbf{y}_i')\}_{i=1}^{l}$ where $\mathbf{y}_i'=F(\mathbf{x}_i')$ for all $i=1,\dots,l$. The posterior $P_Z$ is again a stochastic process with marginals $\psi_{\mathbf{x}_{1:k}}^{P_Z}$ and conditional density given by
\begin{align}
\label{eq:cond_density} 
\lambda_{\mathbf{x}_{1:k}}^{P_Z}(\mathbf{y}_{1:k})=
\lambda_{\mathbf{x}_{1:k}|\mathbf{x}_{1:l}'}^{P}(\mathbf{y}_{1:k}|\mathbf{y}_{1:l}')=\frac{\lambda_{\mathbf{x}_{1:k},\mathbf{x}_{1:l}'}^P(\mathbf{y}_{1:k},\mathbf{y}_{1:l}')}{
\lambda_{\mathbf{x}_{1:l}'}^{P}(\mathbf{y}_{1:l}')
}
\end{align}
\textbf{2. Marginals of transformed process.} 
If $F\sim P$, it holds that $g.P$ has marginals  $\psi_{\mathbf{x}_{1:k}}^{g.P}$ with density given by 
\begin{align}
\label{eq:change_of_variables}
    \lambda_{\mathbf{x}_{1:k}}^{g.P}(\mathbf{y}_{1:k})=\lambda_{g^{-1}\mathbf{x}_{1:k}}^{P}(\rho(h)^{-1}\mathbf{y}_{1:k})
\end{align}
after using a change of variables.\\
\textbf{3. Express invariance in terms of marginals. } By definition, $P$ is $G$-invariant if $g.P=P$ for all $g\in G.$ By Kolmogorov's theorem, this is equivalent to the fact the finite-dimensional marginals of $P$ and $g.P$ agree for all $g\in G$, i.e. 
\begin{align}
    &\psi_{\mathbf{x}_{1:k}}^{P}=\psi_{\mathbf{x}_{1:k}}^{g.P}\tforall \mathbf{x}_1,\dots,\mathbf{x}_k\in\RR^n\\
    \,\Leftrightarrow\,&
         \label{eq:invariant_density}
\lambda_{\mathbf{x}_{1:n}}^P(\mathbf{y}_{1:n})=
    \lambda_{\mathbf{x}_{1:n}}^{g.P}(\mathbf{y}_{1:n})=\lambda_{g^{-1}\mathbf{x}_{1:n}}(\rho(h)^{-1}\mathbf{y}_{1:n})\tforall \mathbf{y}_1,\dots,\mathbf{y}_k\in\mathbb{R}^d,\mathbf{x}_1,\dots,\mathbf{x}_k\in\RR^n
\end{align}
where we used \cref{eq:change_of_variables} in the last equation.\\
\textbf{4. Express equivariance in terms of marginals.} Next, let us be given a context set $Z=\{(\mathbf{x}_i,\mathbf{y}_i)\}_{i=1}^{n}$ where  $\mathbf{y}_i=F(\mathbf{x}_i)$.  We compute:
\begin{align}
    &P_{g.Z}=g.P_{Z} \\
    \Leftrightarrow \,& \,
    \psi^{P_{g.Z}}_{\mathbf{x}_{1:k}}=\psi^{g.P_{Z}}_{\mathbf{x}_{1:k}} \tforall \mathbf{x}_1,\dots,\mathbf{x}_k\in\mathbb{R}^n\\
        \Leftrightarrow \,& \,\lambda^{P_{g.Z}}_{\mathbf{x}_{1:k}}(\mathbf{y}_{1:k})=\lambda^{g.P_{Z}}_{\mathbf{x}_{1:k}}(\mathbf{y}_{1:k}) \tforall \mathbf{x}_1,\dots,\mathbf{x}_k\in\mathbb{R}^n,\mathbf{y}_1,\dots,\mathbf{y}_k\in\mathbb{R}^d\\
        \Leftrightarrow \,& \,\frac{\lambda_{\mathbf{x}_{1:k},g\mathbf{x}_{1:l}'}^P(\mathbf{y}_{1:k},\rho(h)\mathbf{y}_{1:l}')}{
\lambda_{g\mathbf{x}_{1:l}'}^{P}(\rho(h)\mathbf{y}_{1:l}')
}=\lambda_{g^{-1}\mathbf{x}_{1:k}}^{P_Z}(\rho(h)^{-1}\mathbf{y}_{1:k}) \tforall\mathbf{x}_1,\dots,\mathbf{x}_k\in\mathbb{R}^n,\mathbf{y}_1,\dots,\mathbf{y}_k\in\mathbb{R}^d\\
\Leftrightarrow 
\,& \,\label{eq:equivariance_in_marginals}
\frac{\lambda_{\mathbf{x}_{1:k},g\mathbf{x}_{1:l}'}^P(\mathbf{y}_{1:k},\rho(h)\mathbf{y}_{1:l}')}{
\lambda_{g\mathbf{x}_{1:l}'}^{P}(\rho(h)\mathbf{y}_{1:l}')
}=\frac{\lambda_{g^{-1}\mathbf{x}_{1:k},\mathbf{x}_{1:l}'}^P(\rho(h)^{-1}\mathbf{y}_{1:k},\mathbf{y}_{1:l}')}{
\lambda_{\mathbf{x}_{1:l}'}^{P}(\mathbf{y}_{1:l}')
} \tforall\mathbf{x}_1,\dots,\mathbf{x}_k\in\mathbb{R}^n,\mathbf{y}_1,\dots,\mathbf{y}_k\in\mathbb{R}^d
 \end{align}
where we used in row order the following facts: 
\begin{enumerate} 
\item Kolmgorov's theorem.
\item Two distributions coincide if and only if their density coincide (Lebesgue-almost everywhere).
\item \Cref{eq:cond_density} on the left-hand side and \cref{eq:change_of_variables} on the right-hand side.
\item \Cref{eq:cond_density} on the right-hand side.
\end{enumerate}

\textbf{Invariance implies equivariance.}
Assuming $P$ is $G$-invariant, we can use          \cref{eq:invariant_density} to get
\begin{align*}
    \lambda_{\mathbf{x}_{1:k},g\mathbf{x}_{1:l}'}^P(\mathbf{y}_{1:k},\rho(h)\mathbf{y}_{1:l}')=\lambda_{g^{-1}\mathbf{x}_{1:k},\mathbf{x}_{1:l}'}^P(\rho(h)^{-1}\mathbf{y}_{1:k},\mathbf{y}_{1:l}'),\quad \lambda_{g\mathbf{x}_{1:l}'}^{P}(\rho(h)\mathbf{y}_{1:l}')
    =\lambda_{\mathbf{x}_{1:l}'}^{P}(\mathbf{y}_{1:l}')
\end{align*}

Inserting that into the left-hand side of \cref{eq:equivariance_in_marginals}, we see that the equality in \cref{eq:equivariance_in_marginals} is true, i.e. $Z\mapsto P_Z$ is equivariant.

\textbf{Equivariance implies invariance.} By going this computation backward, we can easily show that equivariance implies invariance as well. However, there is a short-cut.  Assuming that $Z\mapsto P_Z$ is equivariant, we can simply pick an empty context set $Z=\{\}$. In this case, $P_{g.Z}=P_{Z}=P$ and therefore equivariance implies $g.P=P$.
\end{proof}
\subsection{Proof of \cref{thm:equivgp}}
\label{proof:equivgp}
\equivgp*
\begin{proof}
A Gaussian process $GP(\mathbf{m},K)$ is $G$-invariant if and only if 
\begin{align*}
    F\sim GP(\mathbf{m},K) \Rightarrow g.F\sim GP(\mathbf{m},K)\text{ for all }g\in G
\end{align*}
By Kolmogorov's theorem (see \citet{KolmogorowReference}), the distribution of $F$ and $g.F$ coincide if and only if their finite-dimensional marginals coincide. Since the marginals are normal, they are equal if and only mean and covariances are equal, i.e. if and only if 
\begin{align}
    \mathbf{m}(\mathbf{x})=&\mathbb{E}(F(\mathbf{x}))
    =\mathbb{E}(g.F(\mathbf{x}))
    =\rho(h)\mathbf{m}(g^{-1}\mathbf{x})
    =g.\mathbf{m}(\mathbf{x})\tforall \mathbf{x}\in\RR^n
\end{align}
and for all $\mathbf{x},\mathbf{x}'\in\RR^n$
\begin{align}
    K(\mathbf{x},\mathbf{x}')=\text{Cov}(F(\mathbf{x}),F(\mathbf{x}'))
    =\text{Cov}(g.F(\mathbf{x}),g.F(\mathbf{x}'))
    =&\text{Cov}(\rho(h)F(g^{-1}\mathbf{x}),\rho(h)F(g^{-1}\mathbf{x}'))\\
    =&\rho(h)\text{Cov}(F(g^{-1}\mathbf{x}),F(g^{-1}\mathbf{x}'))\rho(h)^T\\
    =&\rho(h)K(g^{-1}\mathbf{x},g^{-1}\mathbf{x}')\rho(h)^T
\end{align}
Let us assume that this equation holds. Then picking $g=t_{\mathbf{x}'}$ implies that 
\begin{align*}
    \mathbf{m}(\mathbf{x})&=\mathbf{m}(\mathbf{x}-\mathbf{x}')\\
    K(\mathbf{x},\mathbf{x}')&=K(\mathbf{x}-\mathbf{x}',0)
\end{align*}
i.e. $\mathbf{m}$ is constant and $K$ is stationary. Similiarly, picking $g=h$ implies     \cref{eq:m_invariant_under_rho} and \cref{eq:GP_kernel_constraint}.\\
To prove the opposite direction assuming the constraints from the theorem, we can simply go these computations backwards.
\end{proof}

\subsection{Proof of \cref{prop:SteerCNP}}
\label{proof:SteerCNP}
\SteerCNP*
\begin{proof}
Let $Q_{Z}$ be the output of the model serving as the approximation of posterior distribution $P_{Z}$. It holds $Q_{Z}$ is $G$-equivariant if and only if $Q_{g.Z}=g.Q_{Z}$.\\ 
If $F\sim Q_{Z}$, it holds by standard facts about the normal distribution
\begin{align*}
    g.F(\mathbf{x})=&\rho(h)F(g^{-1}\mathbf{x})\\
    \sim&\mathcal{N}(\rho(h)\mathbf{m}_{Z}(g^{-1}\mathbf{x}),\rho(h)\mathbf{\Sigma}_{Z}(g^{-1}\mathbf{x})\rho(h)^T)\\
    =&\mathcal{N}(g.\mathbf{m}_{Z}(\mathbf{x}),g.\mathbf{\Sigma}_{Z}(\mathbf{x}))
\end{align*}
which gives the one-dimensional marginals of $g.Q_{Z}$. By the conditional independence assumption, $g.Q_{Z}=Q_{g.Z}$ if and only if their one-dimensional marginals agree, i.e. if for all $\mathbf{x}$ 
\begin{align*}
   \mathcal{N}(\mathbf{m}_{g.Z}(\mathbf{x}),\mathbf{\Sigma}_{g.Z}(\mathbf{x}))= \mathcal{N}(g.\mathbf{m}_{Z}(\mathbf{x}),g.\mathbf{\Sigma}_{Z}(\mathbf{x}))
\end{align*}
This is equivalent to $\mathbf{m}_{g.Z}=g.\mathbf{m}_{Z}$ and $\mathbf{\Sigma}_{g.Z}=g.\mathbf{\Sigma}_{Z}$, which finishes the proof.
\end{proof}

\subsection{Proof of \cref{thm:equivdeepsets}}
\label{proof:equivdeepsets}
\equivdeepsets*
\begin{proof}
This proof generalizes the proof of \citet[Theorem 1]{ConvCNP}.\\
\textbf{Step 1: Injectivity of E (up to permutations).}\\ We first want to show that under the given conditions $E$ is injective up to permutations, i.e. $Z=\{(\mathbf{x}_i,\mathbf{y}_i)\}_{i=1}^{m}$ is a permutation of the elements of $Z'=\{(\mathbf{x}_j',\mathbf{y}_j')\}_{j=1}^{m'}$ if and only if $E(Z)=E(Z')$. By definition, $E(Z)=E(Z')$ is equivalent to 
\begin{align}
\label{eq:sum_embedding}
    \sum\limits_{i=1}^{m}K(\cdot,\mathbf{x}_i)\begin{pmatrix}
    1 \\ \mathbf{y}_i
    \end{pmatrix}=  \sum\limits_{j=1}^{m'}K(\cdot,\mathbf{x}_j')\begin{pmatrix}
    1 \\ \mathbf{y}_j'
    \end{pmatrix}
\end{align}
Clearly, if $Z$ is a permutation of $Z'$, \cref{eq:sum_embedding} holds since one can simply change order of summands. Conversely, let us assume that \cref{eq:sum_embedding} holds. Let $f:\Reals^n\to\Reals^d$ be a function in the reproducing kernel Hilbert space (RKHS) of $K$ \citep{Review_matrix_kernels}. The reproducing property in the case of matrix-valued kernels says that
\begin{align}
    \eukl{f}{K(\cdot,\mathbf{x})c}_{\mathcal{H}}=f(\mathbf{x})^Tc \tforall c\in\RR^d,\mathbf{x}\in\RR^n
\end{align}
where $\eukl{\cdot}{\cdot}_{\mathcal{H}}$ is the inner product on the RKHS $\mathcal{H}$. Taking the inner product with $f$ on both sides of \cref{eq:sum_embedding}, we get by the reproducing property:
\begin{align}
    \label{eq:reproducing_property}
    \sum\limits_{i=1}^{m}f(\mathbf{x}_i)^T\begin{pmatrix}
    1 \\ \mathbf{y}_i
    \end{pmatrix}=  \sum\limits_{j=1}^{m'}f(\mathbf{x}_j')^T\begin{pmatrix}
    1 \\ \mathbf{y}_j'
    \end{pmatrix}
\end{align}
Let us choose an arbitrary $\mathbf{x}_k$ where $k=1,\dots,m$ and let us pick $f\in\mathcal{H}$ such that $f(\mathbf{x}_k)=(1,0,\dots,0)^T$,
$f(\mathbf{x}_i)=0$ for all $i\neq k$ and $f(\mathbf{x}_j')=0$ for all $j=1,\dots,m'$ such that $\mathbf{x}_j'\neq \mathbf{x}_k$. This is possible because $K$ is interpolating since we assumed that $K$ is strictly positive definite. In \cref{eq:reproducing_property}, we then get
\begin{align}
    1=\sum\limits_{j=1}^{m'}1_{\mathbf{x}_j'=\mathbf{x}_k}
\end{align}
Therefore, there is exactly one $j$ such that $\mathbf{x}_j'=\mathbf{x}_k$. So every element $\mathbf{x}_k$ from $Z$ can be found exactly once in $Z'$. Turning the argument around by switching $Z$ and $Z'$, we get that also every element $\mathbf{x}_j'$ in $Z'$ can be found exactly once in $Z$. Hence, it holds that $m=m'$ and $(\mathbf{x}_1,\dots,\mathbf{x}_m)$ is a permutation of $(\mathbf{x}_1',\dots,\mathbf{x}_{m}')$. Therefore, we can now assume without loss of generality 
that $\mathbf{x}_i=\mathbf{x}_i'$ for all $i=1,\dots,m$.\\
In \cref{eq:reproducing_property}, pick now $f$ such that $f(\mathbf{x}_i)=(0,\mathbf{y})^T$ for some $\mathbf{y}\in\Reals^d$. Then it follows that 
\begin{align}
    \mathbf{y}^T\mathbf{y}_i=\mathbf{y}^T\mathbf{y}_i'
\end{align}
Since $\mathbf{y}$ was arbitrary, we can conclude that $\mathbf{y}_i=\mathbf{y}_i'$ for all $i=1,\dots,m$. In sum, this shows that $Z$ is a permutation of $Z'$ and concludes the proof that $E$ is injective up to permutations.

\textbf{Step 2: Equivariance of E.}\\
Next, we show that $Z\mapsto E(Z)$ is $G$-equivariant where the transformation of $E(Z)$ is defined by $\rho_E$ as in   \cref{eq:transforming_feature_map}. Let $Z=\{(\mathbf{x}_i,\mathbf{y}_i)\}_{i=1}^{m}$ be a context set and $g=t_{\mathbf{x}}h\in G$. We compute
\begin{align}
    E(g.Z)=\sum\limits_{i=1}^{m}K(\cdot,g\mathbf{x}_i)\begin{pmatrix}
    1 \\ \rho_{\tin}(h)\mathbf{y}_i
    \end{pmatrix}
    =&\sum\limits_{i=1}^{m}K(\cdot,g\mathbf{x}_i)\rho_{E}(h)\begin{pmatrix}
    1 \\ \mathbf{y}_i
    \end{pmatrix}\\
    =&\sum\limits_{i=1}^{m}\rho_{E}(h)K(g^{-1}\cdot,\mathbf{x}_i)\rho_{E}(h)^T\rho_{E}(h)\begin{pmatrix}
    1 \\ \mathbf{y}_i
    \end{pmatrix}\\
    =&\rho_E(h)E(Z)(g^{-1}\cdot)\\
    =&g.E(Z)
\end{align}
where the first equality follows by definition of $E$, the second by definition of $\rho_E$, the third by using $\rho_E$-equivariance of $K$, the fourth by using the assumed orthogonality of $\rho_E$ (see \cref{sec:feature_maps}) and the fifth by definition.

With step 1 and 2, we can now proof the theorem.

\textbf{Step 3: Universality and Equivariance of the decomposition $\Phi=\Psi\circ E$.}

If $\Psi:\mathcal{F}_{\rho_E}\to\mathcal{F}_{\rho_{\tout}}$ is some $G$-equivariant function, it follows that $\Phi=\Psi\circ E$ is $G$-equivariant as well since it is a composition of equivariant maps $\Psi$ and $E$. This shows that the composition is equivariant.

Conversely, if we assume that $\Phi:\mathcal{Z}_{\rho_{\tin}}\to\mathcal{F}_{\rho_{\tout}}$ is a  $G$-equivariant, permutation-invariant function, we can consider it as a function defined on the family  $\mathcal{Z}_{\rho_{\tin}}^{\sim}$ of equivalence classes of sets $Z,Z'\in\mathcal{Z}_{\rho_{\tin}}$ which are permutations of each other. On $\mathcal{Z}_{\rho_{\tin}}^{\sim}$, $E$ is injective and we can define its inverse $E^{-1}$ on the image of $E$ (and set constant zero outside of the image). Clearly, it then holds $\Phi=\Psi\circ E$. Since $E$ is equivariant, also the inverse $E^{-1}$ is and therefore $\Psi$ is equivariant as a composition of equivariant maps $\Phi$ and $E^{-1}$. This shows that this composition is universal.

This finishes the proof of the main statement of the theorem.

\textbf{Additional step: Continuity of $\Phi$.}
We can enforce continuity of $\Phi$ by:
\begin{enumerate}
\item We restrict $\Phi$ on a subset $\mathcal{Z}'\subset\mathcal{Z}_{\rho_{\tin}}$ which is topologically closed, closed under permutations and closed under actions of $G$.
\item $K$ is continuous and $K(\mathbf{x},\mathbf{x}')\to 0$ for $\|\mathbf{x}-\mathbf{x}'\|\to \infty$. 
\item $\Psi:\mathcal{H}\to C_b(\RR^n,\RR^d)$ is continuous, where we denote with $C_b(\RR^n,\RR^d)$  the space of continuous, bounded functions $f:\RR^n\to\RR^d$.
\end{enumerate}

The proof of this follows directly from the proof of the  ConvDeepSets theorem from \citet{ConvCNP}, along with the additional conditions proved above. 
\end{proof}

\section{Divergence-free and Curl-free kernels}

\label{sec:divcurl}
A divergence-free kernel is a matrix-valued kernel $\Phi: \RR^n\times \RR^n \to \RR^{n \times n}$ such that its columns are divergence-free. That is $\nabla^T(\Phi(\x,\x')c) = 0 \; \forall \; c,\x,\x' \in \RR^n$ where the derivatives are taken as a function of $\x$. This ensures that fields constructed by $f(\x) = \sum_{i=1}^N \Phi(\x, \x_i)c_i$ for some $c_i, \x_i \in \RR^n$ are divergence-free. A similar definition holds for curl-free kernels. 

The kernels used in this work were introduced by \citet{macedo2010learning}. In particular we use the curl- and divergence-free kernels with length scale $l>0$ as defined for all $\mathbf{x}_1,\mathbf{x}_2\in\Reals^n$ by
\begin{align}
    K_{\text{curl}}=k_0(\x_1,\x_2)A(\x_1,\x_2),\quad 
    K_{\text{div}}(\x_1,\x_2)=k_0(\x_1,\x_2)B(\x_1,\x_2)
\end{align}
where
\begin{align}
	k_0(\mathbf{x}_1,\mathbf{x}_2)&=\frac{1}{l^{2}}\exp\left(-\frac{\|\mathbf{x}_1-\mathbf{x}_2\|^{2}}{2l^{2}}\right )\\
	A(\mathbf{x}_1,\mathbf{x}_2)&=\mathbf{I}-\frac{(\mathbf{x}_1-\mathbf{x}_2)(\mathbf{x}_1-\mathbf{x}_2)^T}{l^2}
	\\
	B(\mathbf{x}_1,\mathbf{x}_2)&=\frac{(\mathbf{x}_1-\mathbf{x}_2)(\mathbf{x}_1-\mathbf{x}_2)^T}{l^2}+\left( 
	n-1-\frac{\|\mathbf{x}_1-\mathbf{x}_2\|^{2}}{l^{2}}
	\right)\mathbf{I}
\end{align}

To see that $K_{\text{curl}}$ is $E(n)$-equivariant, we compute for $g=t_{\x'}h\in G$
\begin{align}
A(g\x_1,g\x_2)=&\mathbf{I}-\frac{(h\mathbf{x}_1+\x'-h\mathbf{x}_2-\x')(h\mathbf{x}_1+\x'-h\mathbf{x}_2-\x')^T}{l^2}\\
=&hh^{T}-\frac{h(\mathbf{x}_1-\mathbf{x}_2)(\mathbf{x}_1-\mathbf{x}_2)^Th^T}{l^2}\\
=&hA(\x_1,\x_2)h^T
\end{align}
This shows that $K_{\text{curl}}$ is $E(n)$-equivariant since $k_0$ is a $E(n)$-invariant scalar kernel. With a similar computation, one can see that $K_{\text{div}}$ is $E(n)$-equivariant. 

\section{Experimental details}
\label{sec:Experimental details}

For the implementation, we used \emph{PyTorch} \citep{PyTorch}. The github repository for the GP and ERA5 experiments can be found at this \href{https://github.com/PeterHolderrieth/Steerable_CNPs}{link}  
and for the MNIST experiments  \href{https://github.com/MJHutchinson/SteerableCNP}{here}. The models are trained on a mix of GTX 1080, 1080Ti and 2080Ti GPUs.

To set up the SteerCNP model,  we stacked equivariant convolutional layers with NormReLU activation functions in between as a decoder. The smoothing step was performed with a scalar RBF-kernel where the length scale is optimised during training. All hidden layers of the decoder use the regular representation $\rho_\reg$ as a fiber representation $\rho$ of the hidden layers of the decoder if the fiber group $H$ is $C_N$ or $D_N$ and the identity representation $\rho_{\Id}$ for infinite fiber groups. This choice gave the best results and is also consistent with observations in supervised learning problems \citep{E2Steerable}. For every model, we optimised the model architecture independently starting with a number of layers ranging from 3 to 9 and with a number of parameters from $20000$ to $2$ million.
All hyperparameters were optimized by grid search for every model individually and can be found in the afore-mentioned repositories.

For the encoder $E$, we found that the choice of kernels $K$ does not lead to significant differences in performance. Therefore, the results stated here used a diagonal RBF-kernel where we let the length-scale variable as a differentiable parameter. Similar to \citet{ConvCNP}, we found that normalising the last $d$-channels with the first channel improves performance. This operation is clearly invertible and preserves equivariance.

\subsection{GP experiments}
For every sample we have chosen a randomly orientated grid $\mathcal{G}\subset [-10,10]^2$ spread in a circle around the origin and sampled a Gaussian process on it with kernel $K$ with $l=5$. To a set of pairs $\{(\mathbf{x},F(\mathbf{x}))\}_{\mathbf{x}\in\mathcal{G}}$, we add random noise $\epsilon\sim\mathcal{N}(0,\sigma^2)$ with $\sigma=0.05$ on $F(\mathbf{x})$. During training, we randomly split a data set in a context set and in target set. The maximum size of a context set is set to $50$. As usually done for CNPs \citep{CNPs}, the target set includes the context set during training.

\subsection{ERA5 data}
The ERA5 data set consists of weather parameters on a longtitude-latitude grid around the globe. We extracted the data for all points surrounding Memphis, Tennessee, with a distance of less than $520$km giving us approximately $1200$ grid points per weather map.

The weather variables we use are temperature, pressure and wind and we picked hourly data from the winter months December, January and February from years $1980$ to $2018$. Every sample corresponds to one weather map of temperature, pressure and wind in the region at one single point in time. Finally, we split the data set in a training set of $35 000$, a validation set of $17 500$ and test set of $17 500$ weather maps. Similarly, we proceeded for the data set from Southern China. We share the exact pre-processing scripts of the ERA5 data also in our  \href{https://github.com/PeterHolderrieth/Steerable_CNPs}{code}.

\subsection{Image inpainting details}
\label{sec:image_details}

\textbf{MNIST experiments} 

In all the experiments the context sets are drawn from $U(\frac{n_{pixels}}{100}, \frac{n_{pixels}}{2})$. We train with a batchsize of 28. The context points are drawn randomly from each batch and the rest of the pixels used as the target set. We train for 10 epochs using Adam \citep{Adam} with a learning rate of $3\times 10^{-4}$ for all the ConvCNP and SteerCNP models. For the CNP models we train for 30 epochs and use a learning rate of $1\times 10^{-3}$. These values were found using early stopping 
and grid search respectively. Pixel intensities are normalised to lie in the range $[0,1]$.

The dataset is additionally augmented with 10$\%$ blank images (equivalent to adding "no digit" class to the dataset in equal proportion to other classes). The rational behind this is that in the test dataset there are large regions of blank canvas. Given the model is trained on small patches, if we only trained on the MNIST digits the model would encounter these large regions of blank space, which it has never seen before. Including these blank images helped rectify this issue, and empirically led to better performance across the board. The GP lengthscale was optimised over a gird of [0.01,0.05,0.1,0.3,0.5,1.0,2.0,3.0] and the variance the same. The optimal parameters were found to be a length scale of 1.0 and a variance of 0.05.

In addition, we apply a sigmoid function to the mean prediction to ensure the predicted mean in the range $[0,1]$. The covariance activation function is replaced with a softplus and a minimum variance of $0.01$. This keeps the model equivariant as the covariance predicted is now an invariant scalar, rather than an equivariant matrix.

\begin{table}[t]
\small
\centering
\caption{Full results for the MNIST experiments. Mean log-likelihood $\pm$ 1 standard deviation over 3 random model and dataset seeds reported.}
\vspace{0.1in}
\setlength\tabcolsep{3pt}
\begin{tabular}{lrrrrrrrr}
\toprule
Test dataset & \multicolumn{2}{l}{MNIST} & \multicolumn{2}{l}{rotMNIST} & \multicolumn{2}{l}{extrapolate MNIST} & \multicolumn{2}{l}{extrapolate rotMNIST} \\
Train dataset &                   MNIST &                rotMNIST &                   MNIST &                rotMNIST &                   MNIST &                rotMNIST &                   MNIST &                rotMNIST \\
Model              &                         &                         &                         &                         &                         &                         &                         &                         \\
\midrule
GP                 &         {0.39$\pm$0.30} &         {0.39$\pm$0.30} &         {0.49$\pm$0.51} &         {0.49$\pm$0.51} &         {0.65$\pm$0.20} &         {0.65$\pm$0.20} &         {0.72$\pm$0.17} &         {0.72$\pm$0.17} \\
CNP                &         {0.76$\pm$0.05} &         {0.66$\pm$0.06} &         {0.53$\pm$0.04} &         {0.69$\pm$0.06} &        {-1.20$\pm$0.06} &        {-1.04$\pm$0.24} &        {-1.11$\pm$0.06} &        {-0.96$\pm$0.22} \\
ConvCNP            &         {1.01$\pm$0.01} &         {0.95$\pm$0.01} &         {0.93$\pm$0.05} &         {1.00$\pm$0.05} &         {1.09$\pm$0.03} &  \textbf{1.11$\pm$0.04} &         {1.08$\pm$0.02} &         {1.14$\pm$0.03} \\
SteerCNP($C_4$)    &         {1.05$\pm$0.02} &  \textbf{1.02$\pm$0.03} &  \textbf{1.01$\pm$0.04} &  \textbf{1.06$\pm$0.04} &         {1.12$\pm$0.02} &  \textbf{1.13$\pm$0.03} &         {1.14$\pm$0.02} &  \textbf{1.16$\pm$0.04} \\
SteerCNP($C_{8}$)  &  \textbf{1.07$\pm$0.03} &  \textbf{1.05$\pm$0.04} &  \textbf{1.04$\pm$0.03} &  \textbf{1.09$\pm$0.03} &         {1.13$\pm$0.01} &  \textbf{1.14$\pm$0.02} &  \textbf{1.16$\pm$0.03} &  \textbf{1.18$\pm$0.02} \\
SteerCNP($C_{16}$) &  \textbf{1.08$\pm$0.03} &  \textbf{1.04$\pm$0.03} &  \textbf{1.04$\pm$0.08} &  \textbf{1.09$\pm$0.07} &  \textbf{1.14$\pm$0.04} &  \textbf{1.11$\pm$0.08} &  \textbf{1.17$\pm$0.05} &  \textbf{1.15$\pm$0.06} \\
SteerCNP($D_4$)    &  \textbf{1.08$\pm$0.03} &  \textbf{1.05$\pm$0.03} &  \textbf{1.04$\pm$0.01} &  \textbf{1.09$\pm$0.03} &  \textbf{1.12$\pm$0.05} &  \textbf{1.13$\pm$0.04} &         {1.14$\pm$0.03} &  \textbf{1.17$\pm$0.06} \\
SteerCNP($D_8$)    &  \textbf{1.08$\pm$0.03} &  \textbf{1.04$\pm$0.04} &  \textbf{1.03$\pm$0.11} &  \textbf{1.10$\pm$0.06} &  \textbf{1.15$\pm$0.02} &  \textbf{1.12$\pm$0.02} &  \textbf{1.17$\pm$0.02} &  \textbf{1.17$\pm$0.02} \\
\bottomrule
\end{tabular}

\label{tab:mnist}
\end{table}

\begin{figure}[h]
    \centering
    \includegraphics[width=\textwidth]{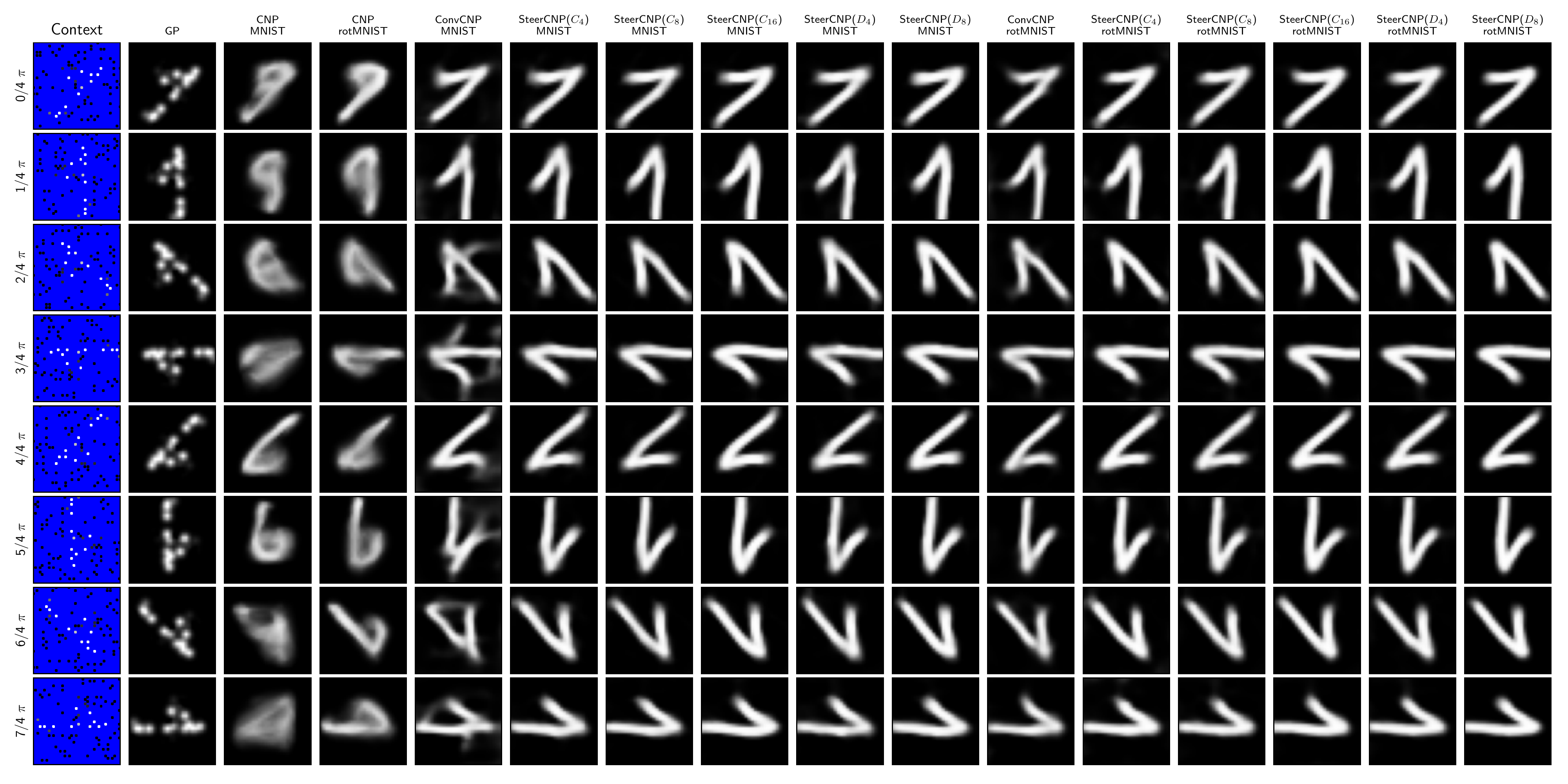}
    \caption{Qualitative examples of the behaviour of the predicted mean of different models when the context set is rotated. We observe that the ConvCNP, even when trained on rotation augmented data, has trouble predicting good shapes when the context set is rotated. By comparison the equivariant models have very consistent predictions under rotation, with the $C_4$ and $D_4$ models beng exactly equivariant to $90^\circ$ rotations, and the $C_{16}$ models being exactly equivariant to $22.5^\circ$ rotations}
\end{figure}

\begin{figure}[h]
    \centering
    \includegraphics[width=\textwidth]{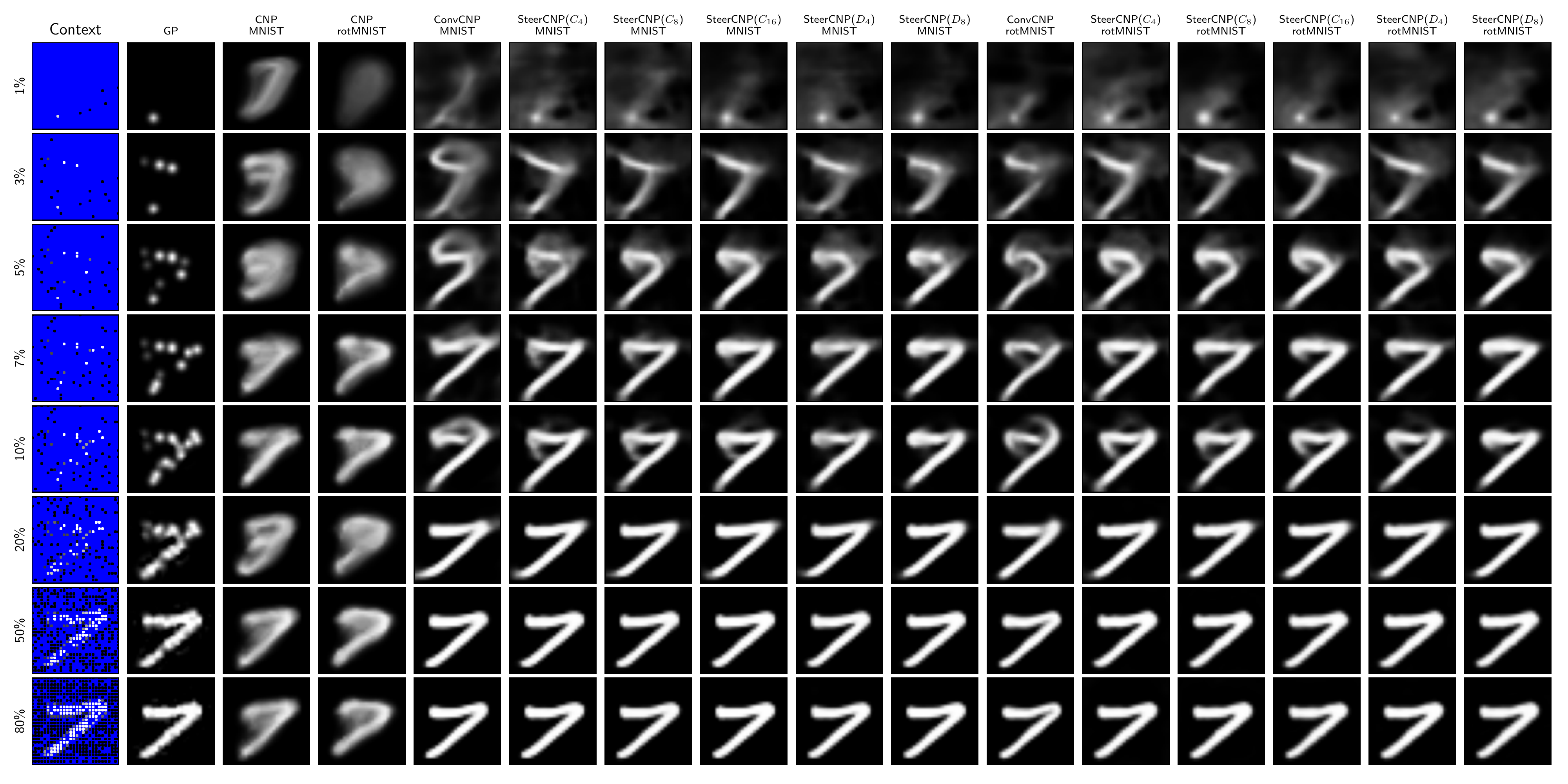}
    \caption{Qualitative examples of the behaviour of the predicted mean of different models when the context set size is changed. Digit not rotated.}
    \centering
    \includegraphics[width=\textwidth]{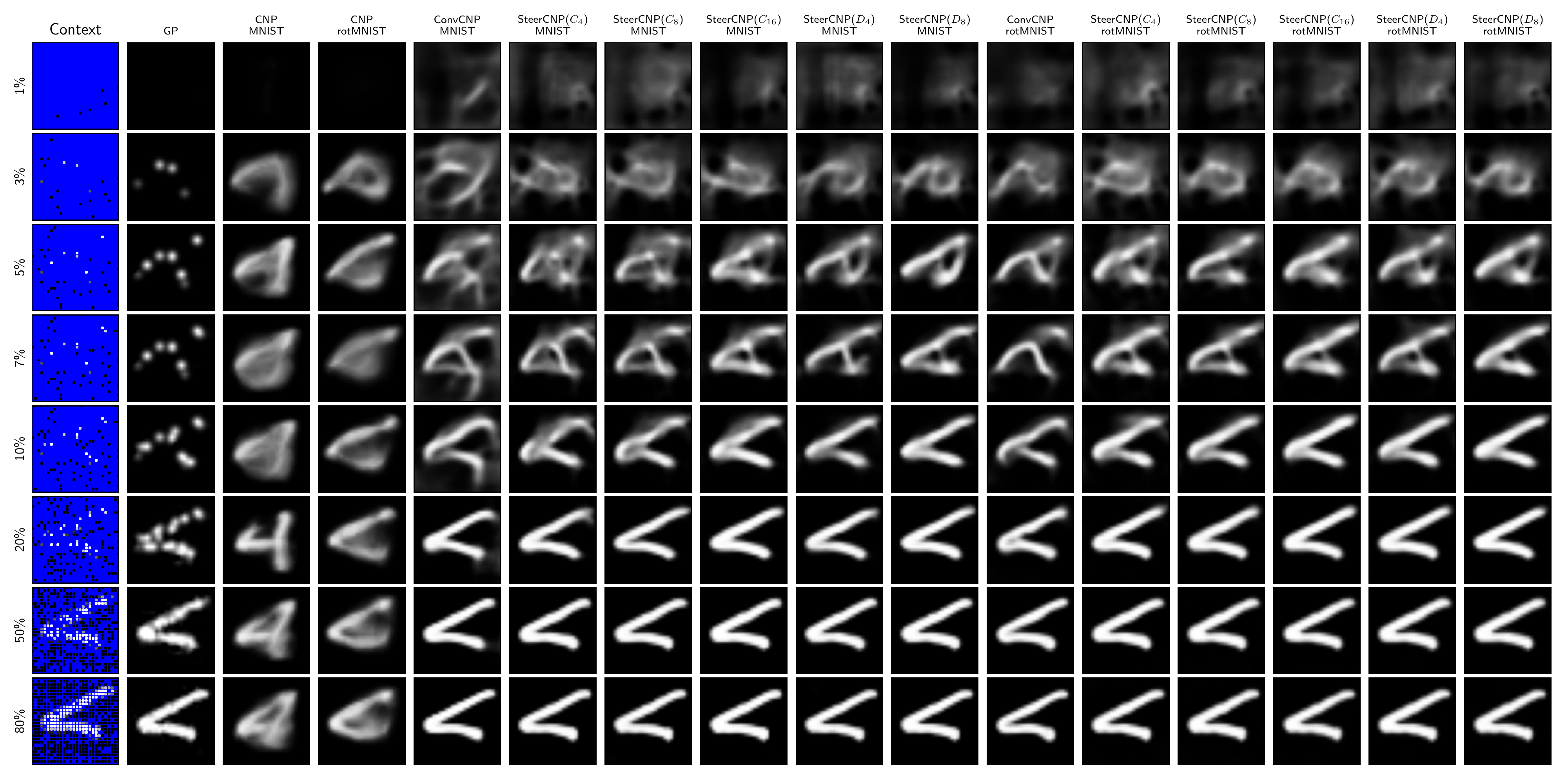}
    \caption{Qualitative examples of the behaviour of the predicted mean of different models when the context set size is changed. Digit rotated.}
\end{figure}

\begin{figure}[h]
    \centering
    \includegraphics[width=\textwidth]{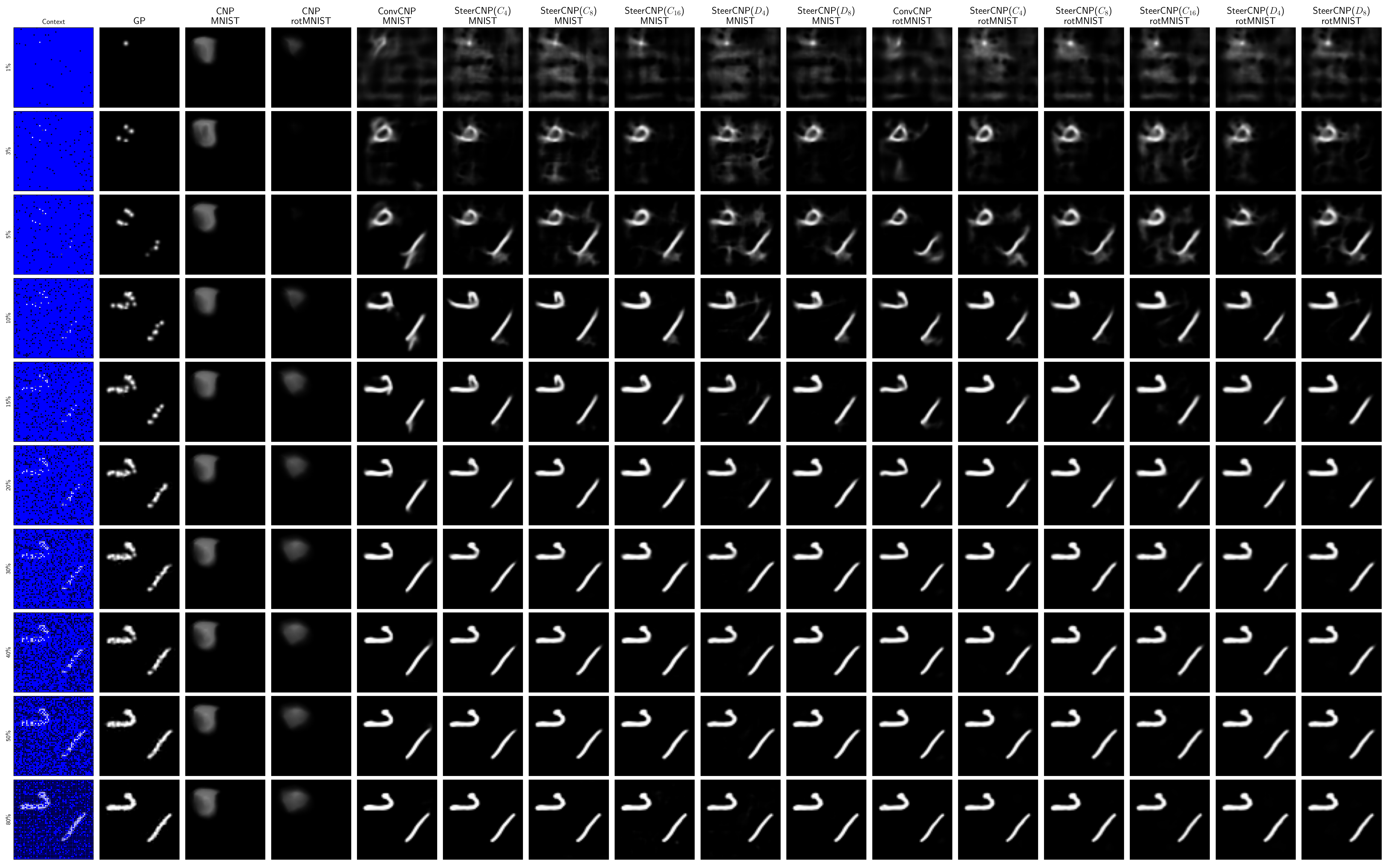}
    \caption{Qualitative examples of the behaviour of models trained on single MNIST digits, tested on multiple digits pasted into a larger canvas. Size of context set varied. We see that further away from the digits there is some noise predictions. These are likely causes by the models never having seen data as far from digits as this, leading to somewhat undefined behaviour. We see that the equivariant models exhibit considerably less of this noisy behaviour.}
\end{figure}

\begin{figure}
    \centering
    \includegraphics[width=\textwidth]{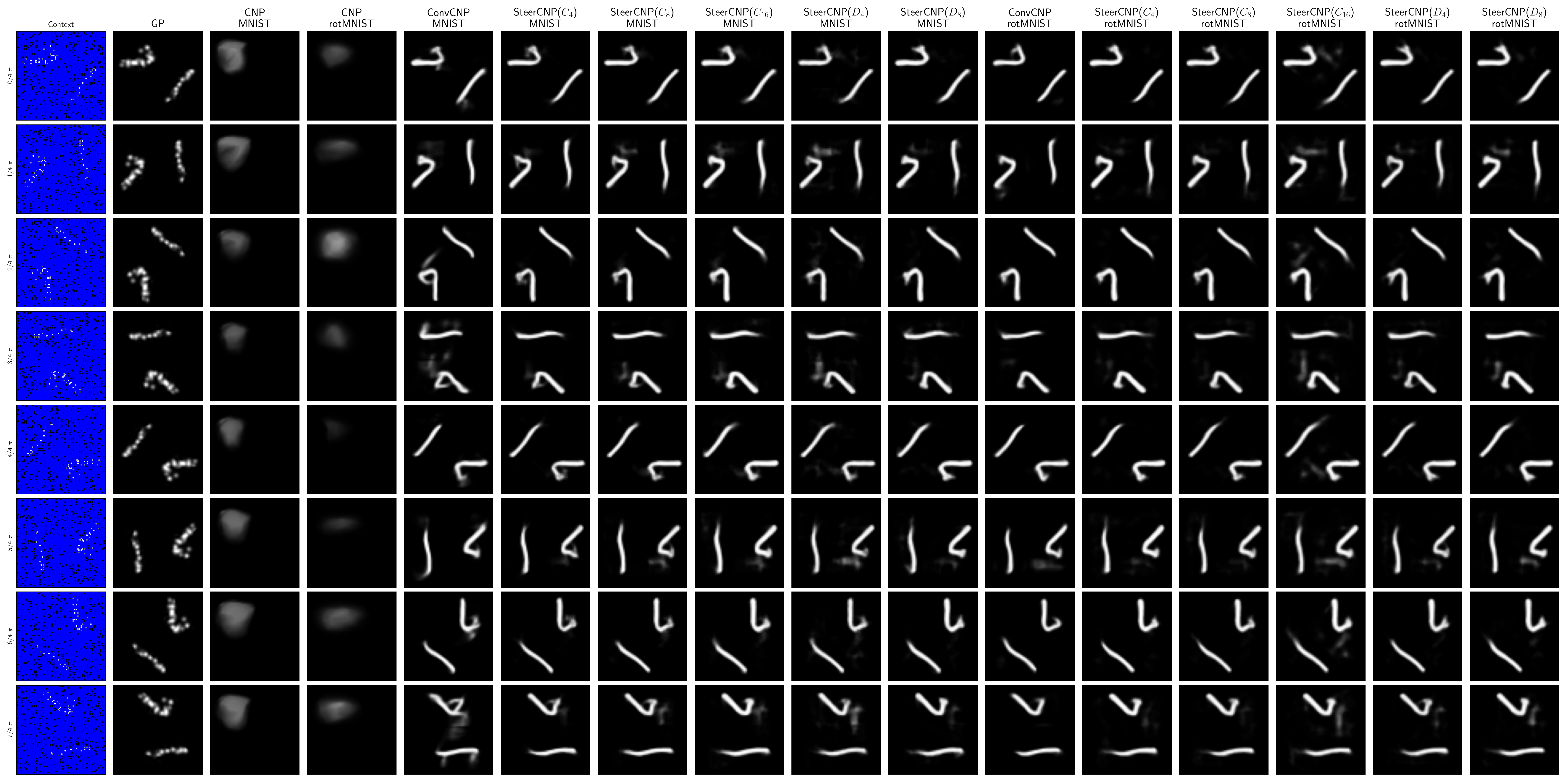}
    \caption{Qualitative examples of the behaviour of models trained on single MNIST digits, tested on multiple digits pasted into a larger canvas. Rotation of the context set is varied. We see that the equivariant models show very little change in behaviour under rotation, whereas the ConvCNP gives reasonably wild predictions under rotation.}
\end{figure}
\end{document}